%% file: main.tex
\documentclass[10pt,journal,compsoc]{IEEEtran}

\input{_preamble}

\begin{document}

\title{Foundational Models for 3D Point Clouds: \\A Survey and Outlook}

\author{
    Vishal Thengane,
    Xiatian Zhu,
    Salim Bouzerdoum,
    Son Lam Phung,
    Yunpeng Li

    \IEEEcompsocitemizethanks{\IEEEcompsocthanksitem  V. Thengane and X. Zhu are with the University of Surrey, United Kingdom.
    \IEEEcompsocthanksitem S. Bouzerdoum and S. L. Phung are with the University of Wollongong, Australia.
     \IEEEcompsocthanksitem Y. Li is with King's College London, United Kingdom.
    \IEEEcompsocthanksitem Correspondence e-mail: v.thengane@surrey.ac.uk
    }
    
}


\input{_sections/0_abstract}

\maketitle

\IEEEdisplaynontitleabstractindextext

\IEEEpeerreviewmaketitle

\input{_sections/1_introduction}
\input{_sections/2_background}

\input{_sections/3_3d_foundation}
\input{_sections/4_2d_vlms_for_3d}
\input{_sections/5_2d_vlms+llms+for_3d}
\input{_sections/6_outlook_into_future}

\input{_sections/7_conclusion}





\begin{small}
    \bibliographystyle{IEEEtran}
    \bibliography{main}
\end{small}

\end{document}

%% file: _preamble.tex
\usepackage{graphicx}            
\usepackage{amsmath, amssymb, amsfonts}  
\usepackage{enumitem}            
\usepackage{booktabs}            
\usepackage{multirow}            
\usepackage{tabularx}            
\usepackage{array}               

\usepackage[table]{xcolor}      
\usepackage[most]{tcolorbox}    
\usepackage{adjustbox}           

\usepackage[
    colorlinks=true,       
    linktocpage=true,      
    backref=page,          
    linkcolor=myred,        
    citecolor=myblue,       
    urlcolor=myred          
]{hyperref}

\usepackage{tikz}                
\usepackage{overpic,picture}     

\usepackage[numbers, square, comma]{natbib}  
\setcitestyle{open={[}, close={]}, separator={,}} 

\definecolor{myblue}{HTML}{7F7FFF} 
\definecolor{myred}{HTML}{E95332}
\definecolor{mygreen}{HTML}{199524}

\newcommand{\tss}[1]{\textsuperscript{#1}}
\newcommand{\tbf}[1]{\textbf{#1}}

\ifCLASSINFOpdf
\else
\fi



%% file: _sections/0_abstract.tex
\IEEEtitleabstractindextext{%
\begin{abstract}

The 3D point cloud representation plays a crucial role in preserving the geometric fidelity of the physical world, enabling more accurate understanding and interaction with complex 3D environments.
While humans naturally comprehend the intricate relationships between objects, their spatial arrangements, and variations through a multisensory system, artificial intelligence (AI) systems have yet to fully replicate this capacity. To bridge this gap, it becomes essential to incorporate multiple modalities, such as images, text, audio, and point clouds. Models that can seamlessly integrate and reason across these modalities are known as foundation models (FMs). The development of FMs for 2D modalities, such as images and text, has seen significant progress, driven by the abundant availability of large-scale datasets. However, the 3D domain has lagged due to the scarcity of labelled data and high computational overheads. In response, recent research has begun to explore the potential of applying FMs to 3D tasks, overcoming these challenges by leveraging existing 2D knowledge. 
Additionally, language, with its capacity for abstract reasoning and description of the environment, offers a promising avenue for enhancing 3D understanding through large pre-trained language models (LLMs). 
Despite the rapid development and adoption of FMs for 3D vision tasks in recent years, there remains a gap in comprehensive and in-depth literature reviews. This article aims to address this gap by presenting a comprehensive overview of the state-of-the-art methods that utilize FMs for 3D visual understanding. 
We start by reviewing various strategies employed in the building of various 3D FMs. Then we categorize and summarize use of different FMs for tasks such as perception tasks. Finally, the article offers insights into future directions for research and development in this field. This survey is intended to serve as a structured guide for researchers and practitioners seeking to delve into this emerging area of study, providing both a summary of existing knowledge and a roadmap for future exploration. To complement this survey, we provide a curated list of 
relevant papers on the topic: \href{https://github.com/vgthengane/Awesome-FMs-in-3D}{https://github.com/vgthengane/Awesome-FMs-in-3D}

\end{abstract}

}

%% file: _sections/1_introduction.tex
\IEEEraisesectionheading{
\section{Introduction}
\label{sec:introduction}}
In the ongoing race to develop artificial intelligence (AI) systems that think and behave like humans, a crucial factor is the ability to navigate and understand the three-dimensional (3D) world around us. For AI systems to be effectively deployed in real-world environments, they must possess a robust sense of 3D world \cite{heaven2016ai3d}. The 3D world can be represented in various formats, including depth images, meshes, volumetric grids, and point clouds \cite{lahoud20223d}. Among these, point clouds are commonly used, consisting of a collection of points in a 3D coordinate system \cite{wiki_pcloud}. 

3D point clouds represent a fundamental paradigm in the domain of spatial data representation \cite{qi2017pointnet}. They serve as a pivotal data structure in several fields, including computer vision, robotics, autonomous vehicles, augmented reality, and many more \cite{lu2022transformers}. In computer vision, point clouds enable the precise modelling of real-world scenes, facilitating tasks such as object detection, scene understanding, and 3D reconstruction \cite{lahoud20223d}. Similarly, point clouds play a crucial role in perception and navigation in robotics and autonomous vehicles, aiding in obstacle detection, environment mapping, and path planning \cite{mao20233d}. Furthermore, in augmented reality applications, point clouds serve as the backbone to overlay virtual objects on a physical world, enhancing user experiences and interactions \cite{fei2023self}. Overall, the versatility and richness of information encapsulated within point clouds make them indispensable tools for 3D understanding and interaction. 

Despite the crucial role of point clouds, their utilisation poses significant challenges.
First, the process of collecting 3D datasets incurs substantial costs and time investments due to the complexities involved in 
capturing spatial information \cite{dai2017scannet}. Additionally, annotating point cloud data with ground truth labels for tasks such as object recognition, semantic segmentation, and reasoning is labour-intensive and requires specialised expertise \cite{yeshwanth2023scannet++}. Training large-scale models on extensive datasets demands considerable computational resources and infrastructure, often necessitating high-performance computing systems \cite{schult2023mask3d}. Moreover, point cloud datasets are inherently sparse, lacking semantic information about objects or scenes, despite capturing geometric details \cite{wang2023unibev}. 

All these challenges have prompted researchers to explore crucial questions: \textit{Can we leverage other available data modalities, such as images, text, and audio, to enhance understanding of 3D data, using models that can extract their features? Furthermore, can we bridge the gaps of limited data, annotations, and semantic information without the need for extensive data acquisition and costly model training?}

This inquiry led to foundational models (FMs). The term ``Foundation Model'', introduced in \cite{bommasani2021opportunities}, refers to deep learning models trained on extensive datasets, often using self-supervision at scale. These models exhibit unprecedented adaptability across diverse tasks and domains, characterised by \textit{pre-training} \cite{hinton2006fast}, \textit{generalizability}, \textit{adaptability} through transfer learning \cite{perkins1999transfer}, \textit{large} scale in both model and data dimensions, and \textit{self-supervised learning} nature. 

Although the fundamental components of foundation models (FMs), such as neural networks and transfer learning, have existed for many years, their significant advancements have recently surged in natural language processing (NLP) with the emergence of large language models (LLMs) such as BERT and GPT-3 \cite{devlin2018bert, brown2020language}. Following their success in NLP, similar progress has been made in computer vision (CV). Vision-language models (VLMs), such as CLIP \cite{radford2021learning}, exemplify this trend, trained on large image-text datasets, demonstrates remarkable generalisability across various downstream tasks \cite{zhang2022pointclipv2, thengane2022clip}. Models like SAM \cite{kirillov2023segment} for segmentation tasks further extend this concept to applications such as class-agnostic segmentation, which can be adapted to various domains, including medical image segmentation \cite{zhang2023segment} and 3D vision \cite{yang2023sam3d}.
 
In the quest to comprehend 3D worlds more effectively, leveraging additional modalities such as images, text, and audio alongside FMs has driven the development of various methods \footnote{For clarity, we will refer to pre-trained models based solely on text as LLMs and those incorporating modalities such as image, audio, video, and text as 2DFMs.}. 
For example, one line of approaches involves constructing 3DFMs by utilising 2DFMs \cite{xue2023ulip, xue2023ulip2}. Another avenue of research uses these 2DFMs for tasks, including point cloud classification \cite{zhang2022pointclip, zhang2022pointclipv2}, tasks such as segmentation \cite{liu2023partslip, umam2023partdistill}, and object detection \cite{bai2022transfusion, liu2023bevfusion}. 
Furthermore, with the advent of LLMs \cite{touvron2023llama, vicuna2023, gunasekar2023textbooks} becoming open source, several methodologies have emerged that use these models for 3D understanding at both object \cite{luo2024scalable, qi2025shapellm} and scene levels \cite{zhou2023regionblip, chen2024ll3da}. 
Although LLMs are inherently designed for text-based reasoning, they are adapted for 3D tasks through integration with vision-based models. For instance, embeddings generated by LLMs from textual descriptions or instructions can be aligned with features from 3D models, enabling tasks such as visual grounding \cite{roh2022languagerefer}, 3D captioning \cite{huang2023embodied}, and 3D question-answering \cite{9965773}. 

Despite the rapid evolution and widespread adoption of 2DFMs in the realm of 3D vision tasks, the literature lacks an in-depth summary of available methodologies. To address this gap, we present a comprehensive and structured guide, with the aim of providing a definitive resource for researchers and practitioners alike.

\input{_sections/_misc/fig_taxonomy.tex}
\textbf{Taxonomy}
This survey presents a detailed analysis of 2DFMs tailored for 3D point cloud understanding.
Designed to benefit researchers including those new and experienced in the field, it offers a structured taxonomy to navigate essential concepts, illustrated in Fig. \ref{fig:taxonomy}. We begin by laying a solid foundation through a detailed discussion of key topics, including point clouds, available datasets, uni-modal and multi-modal models, and downstream adaptation. Sec. \ref{sec:building_3dfms} then delves into various initial efforts to build 3DFMs, using 2DFMs. Subsequently, in Sec. \ref{sec:perception_tasks}, we examine the applications of these 2D- and 3DFMs to address various 3D tasks such as classification, segmentation, and detection.
Similarly, Sec. \ref{sec:reasoning_tasks} explores the application of 2D- and 3DFMs together with LLMs for 3D tasks. 
Throughout, we summarise methods and provide insights into their performance on different datasets used, 
Sec. \ref{sec:outlook_into_future} offers an outlook on current limitations and future directions, concluding our review in Sec. \ref{sec:conclusion}. 

\textbf{Scope} Our survey focusses on examining FMs specifically for 3D point clouds.
These FMs encompass both uni-modal models tailored for texts, commonly known as LLMs, such as LLaMa \cite{touvron2023llama}, GPT-3 \cite{brown2020language}, and Vacuna \cite{vicuna2023}, as well as their multi-modal counterparts such as CLIP \cite{radford2021learning}, SAM \cite{kirillov2023segment}, ImageBind \cite{han2023imagebind} and their variants \cite{jia2021scaling} and variants of LLMs such as LLaVa \cite{liu2024visual}, MiniGPT-4 \cite{zhu2023minigpt}. 
We exclude methods that use 2DFMs for tasks such as generation, manipulation, or rendering, as they have been extensively covered in the existing literature. Furthermore, we do not delve into applications within specific domains such as medical imaging or remote sensing, as these warrant dedicated survey papers. Instead, we offer a broader overview of the grounding work present in the literature, which can be readily applied across various domains.

\textbf{Related surveys} We compare this work with existing surveys in the 3D literature. Guo et al. \cite{guo2020deep} present a comprehensive review of deep learning methods for 3D point clouds. Similarly, \cite{lu2022transformers, lahoud20223d, zeng2022survey} provide detailed analyses but focus exclusively on models based on transformer architectures. Several works summarise methods for 3D object detection in autonomous driving only \cite{mao20233d, wang2023multi, prakash2021multi, tang2023multi} and do not cover general use cases. Additionally, these works are outdated, as there have been significant advancements utilising pre-trained large models for 3D understanding. Awaise et al. \cite{awais2023foundational} summarise 2DFMs for computer vision tasks, but do not review 3D use cases. Other works such as \cite{fei2023self, xiao2023survey} have limited scopes; for instance, \cite{fei2023self} only provides a summary of self-supervised methods for point clouds, and \cite{xiao2023survey} focusses on label-efficient methods for point clouds. In contrast, this survey aims to provide a comprehensive list of methods, to the best of our knowledge, that utilise 2D/3D FMs to solve various 3D downstream tasks.

\textbf{Features}
This survey represents the first comprehensive exploration of the landscape of FMs for 3D point-cloud learning, filling a significant gap in the current literature and intended to serve as a starting guide for both newcomers and seasoned researchers in the domain. The key features of this survey include:

\begin{itemize}[label={\color{myblue}{$\blacktriangleright$}}, left=2pt]
    \item \textit{Background on 3D Vision Tasks and Datasets}: Essential background on point clouds and an overview of diverse datasets for training and evaluation, highlighting their key characteristics and challenges.

    \item \textit{Discussion of FMs and Key Concepts}: A concise explanation of FMs and essential terminology to ensure clarity in understanding their applications across various use cases.

    \item \textit{Comprehensive Analysis of Methods}: A detailed review of existing methods, including comparisons with alternatives, provides the reader with a clear understanding of their strengths, limitations, and applications.

\end{itemize}

\textbf{Contributions}
The contributions of this work are as follows:
\begin{itemize}[label={\color{myblue}{$\blacktriangleright$}}, left=2pt]
    \item \textit{Comprehensive Background Explanation}: We provide an introduction about point clouds, various available datasets for 3D point cloud understanding, and the FMs and important terminology. This background sets the stage for understanding the methodologies discussed in this survey.
    
    \item \textit{Structured Taxonomy}: We present a structured taxonomy that offers clarity and ease of understanding for both new researchers and those seeking deeper insights into current trends in the field. Our taxonomy groups methods based on different tasks, adaptation strategies, and other important factors, facilitating better organisation and comprehension of the surveyed literature.
    
    \item \textit{Insightful Discussion on Future Prospects}: In addition, we offer an insightful discussion on future prospects based on the papers discussed in this work. This discussion covers aspects related to datasets, effective methods for adapting these models for 3D tasks, and other emerging trends in the field. 
\end{itemize}

By offering a comprehensive synthesis of FMs, taxonomy, datasets, and methodologies, this survey serves as a valuable guide for researchers, practitioners, and enthusiasts alike, with the aim of advancing the field of 3D world understanding.

%% file: _sections/_misc/fig_taxonomy.tex
\begin{figure}[!htb]
  \centering
  \includegraphics[width=\linewidth]{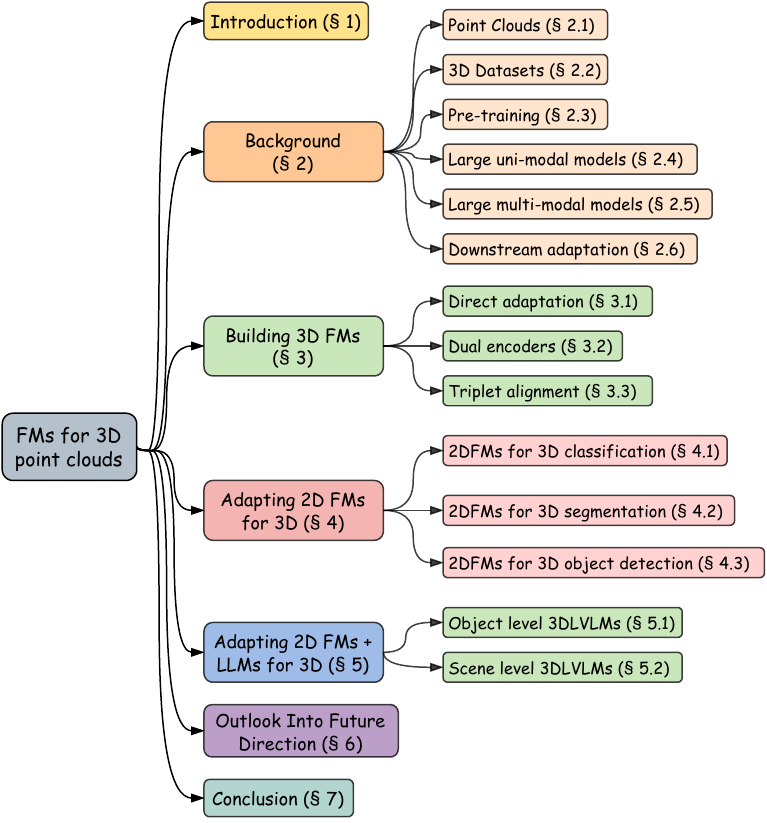}
  \caption{Taxonomy of foundational models for 3D point clouds.}
  \label{fig:taxonomy}
\end{figure}

%% file: _sections/2_background.tex
\section{Background}
\label{sec:background}

\begin{figure}[!htb]
  \centering
  \includegraphics[width=\linewidth]{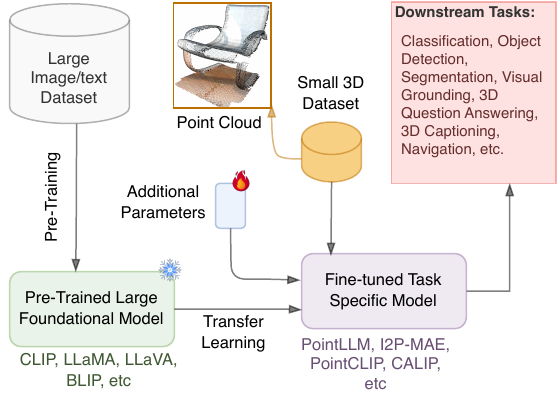}
  \caption{A general pipeline of using FMs for 3D vision tasks.}
  \label{fig:background_ills}
\end{figure}

This section serves as an introductory overview, designed to provide readers with essential background and familiarise beginners with key concepts necessary to understand the context of this survey. 
First, we explain what 3D point clouds are and how they're represented (Sec. \ref{sec:bg_pcloud}). Following this, 
we will briefly introduce the available 3D point cloud datasets, including object-level, scene-level, and their extended versions 
(Sec. \ref{sec:bg_datasets}). 
Next, we will explain the pre-training technique, emphasising its significance in enhancing model performance (Sec. \ref{sec:bg_pretrain}). Then, we introduce large uni-modal  multi-modal models (Sec. \ref{sec:bg_multimodal}), focussing on their applications, impact, and relevance to 3D point cloud learning. Finally, we discuss the downstream adoption technique, which plays a critical role in adapting pre-trained models to new tasks (Sec. \ref{sec:bg_adaptation}). 
We refer the readers to Fig. \ref{fig:background_ills} for an illustration of the interconnections of the topics covered in this section.

\subsection{Point Clouds}
\label{sec:bg_pcloud}

3D data can take on diverse forms, including multi-view images, meshes, volumetric displays, voxel grids, depth maps, and notably point clouds. Point clouds emerge as particularly prevalent representations for 3D data. A point cloud, denoted as $\mathbf{P}$, constitutes a compilation of 3D vectors (points), expressed as $\mathbf{P} = \{p_1, p_2, p_3, \ldots, p_N\}$, where each vector is defined as a point $p_i = \{\mathbf{C}_i, \mathbf{F}_i\}$. In this context, $\mathbf{C}_i \in \mathbb{R}^{1 \times 3}$ signifies the 3D coordinates $(x_i, y_i, z_i)$ of the point, and $\mathbf{F}_i$ encompasses the attributes of the characteristic of the point such as RGB values, intensity, normal vectors, etc. These attributes are discretionary and subject to variation depending on factors such as the 3D sensor used, the data collection process, and specific application requirements. This data is often captured using 3D scanners like LiDAR or RGB-D sensors, potentially incorporating colour information from RGB cameras. Notably distinct from images represented as matrices, point clouds are unordered sets. Therefore, processing such data necessitates a permutation-invariant approach to ensure consistent output, regardless of the ordering of the same point cloud. The adaptability and versatility of point clouds make them indispensable for capturing and processing 3D spatial information across diverse applications.

\subsection{3D Datasets}
\label{sec:bg_datasets}
Here, we summarise commonly used datasets for point cloud understanding tasks such as classification, detection, and segmentation, along with recently developed multi-modality 3D datasets that address the data requirements of large models. We categorise these datasets into three groups: object-level, scene-level, and their extensions for 3D language understanding tasks.

\subsubsection{Object Level Datasets}
Building a large-scale, realistic 3D database is both costly and challenging. 
An approach is to leverage controllable data synthesis.
For instance, ShapeNet \cite{chang_shapenet_2015} offers 51,300 CAD models across 55 categories, while ModelNet40 \cite{Wu_2015_CVPR} includes 12,311 models spanning 40 categories. Additionally, datasets such as 3D-FUTURE \cite{fu20213d} and ABO \cite{collins2022abo} enhance this collection with CAD models that feature intricate geometry and detailed textures. However, the gap between synthetic and real-world objects continues to drive the demand for extensive, real-world 3D object datasets. To this end, DTU \cite{Jensen_2014_CVPR} and BlendedMVS \cite{yao2020blendedmvs} provide photorealistic data intended for multi-view stereo benchmarks. ScanObjectNN \cite{uy2029scanobjectnn}, derived from scanned indoor environments, includes approximately 15,000 coloured point cloud objects across 15 categories. GSO \cite{downs2022google} offers 1,030 scanned objects with fine-grained geometry and texture for 17 common household items, while AKB-48 \cite{liu2022akb}, focused on robotics, provides 2,037 articulated models across 48 categories. CO3D \cite{reizenstein2021common} presents 40,000 object-centric video clips with annotated point clouds generated by COLMAP. Finally, the OmniObject3D \cite{wu2023omniobject3d} dataset comprises 6,000 3D objects, including meshes, textures, and multi-view images across 190 everyday categories.

\subsubsection{Scene Level Datasets}
Scene-level datasets are essential for a wide range of applications in 3D domain. Several datasets have emerged for both indoor and outdoor environments. For example, Matterport3D \cite{Matterport3D} consists of low-resolution reconstructions derived from panoramic RGB-D images and includes semantic annotations. RealEstate10k \cite{zhou2018stereo} features camera poses corresponding to 10 million frames extracted from approximately 80,000 video clips gathered from around 10,000 YouTube videos. ARKitScenes \cite{baruch2021arkitscenes} enhances ground-truth geometry by providing box annotations for 17 object classes. ScanNet \cite{dai2017scannet} offers 3D reconstructions and annotations at scale, comprising 1,503 RGB-D sequences from 707 unique scenes recorded using an iPad mounted with a structure sensor. Building on the ScanNet dataset, ScanNet200 \cite{rozenberszki2022language} focusses on the recognition of 200 annotated classes, while ScanNet++ \cite{yeshwanth2023scannet++} expands this with 460 scenes, 280,000 captured DSLR images, and over 3.7 million iPhone RGB-D frames.

In outdoor domain, SemanticKITTI \cite{behley2019iccv} provides dense point-wise annotations for the complete 360-degree field of view captured by automotive LiDAR. Paris-CARLA-3D \cite{deschaud2021paris} supports semantic, instance, and scene completion tasks in dense point clouds for outdoor mapping. nuScenes \cite{caesar2020nuscenes} features a comprehensive suite of autonomous vehicle sensors, including 6 cameras, 5 radars, and 1 LiDAR, all with a 360-degree field of view. It comprises 1,000 scenes, each 20 seconds long, fully annotated with 3D bounding boxes for 23 classes and 8 attributes.

\subsubsection{Modified 3D Datasets}
The datasets highlighted above do not incorporate a understanding of textual information. To effectively employ large language models for 3D understanding, it is necessary to develop 3D-language or multimodal datasets. These datasets are built upon earlier datasets such as ScanNet \cite{dai_scannet_2017}, Matterport3D \cite{Matterport3D}, ShapeNet \cite{chang_shapenet_2015}, and nuScenes \cite{caesar_nuscenes_2020}. In the following, we highlight datasets that have been developed on top of these foundations.

Cap3D \cite{luo2024scalable} utilises 660,000 objects from Objaverse \cite{deitke2023objaverse} to consolidate multi-view 2D image captions of 3D objects. Text2Shape \cite{chen2018text2shape} leverages human-annotated data from ShapeNet, creating templates for generative text-to-3D tasks. SceneVerse \cite{jia2024sceneverse} compiles 68,000 annotated scenes with 2.5 million vision-language pairs generated through 3D scene graphs and LLMs to support object and scene captioning. nu-Caption \cite{yang2023lidar} annotates 420,000 LiDAR scans from nuScenes using GPT-4 \cite{achiam2023gpt} to provide scene descriptions and object relationships. Building on nu-Caption, nu-Grounding \cite{yang2023lidar} facilitates visual grounding tasks with 280,000 question-answer pairs. ScanRefer \cite{chen2020scanrefer} introduces natural language grounding for 3D scenes, featuring over 51,000 annotated referring expressions in 800 ScanNet scenes. ReferIt3D \cite{achlioptas2020referit_3d} refines this task by focussing on disambiguation in scenes with multiple object instances. Lastly, Multi3DRefer \cite{zhang2023multi3drefer} extends ScanRefer by providing referring descriptions for zero-target, single-target, and multiple-target, which accommodate more complex queries.

For a more exhaustive review of such extended datasets, please refer to \cite{ma2024llms}, which summarises an extensive range of datasets in the 3D vision-language domain.

\subsection{Pre-Training}
\label{sec:bg_pretrain}
The concept of pre-training \cite{donahue2014decaf} has been widely used since the early days of computer vision, particularly with the advent of large datasets like ImageNet \cite{russakovsky2015imagenet}. When substantial data availability and computational resources are accessible, pre-training involves training a large model on an extensive dataset using supervised, unsupervised, self-supervised, or weakly supervised methods, as evidenced by significant work in the field \cite{he2016deep, radford2021learning, jia2021scaling, kirillov2023segment}. The main idea is that the model learns diverse features, enabling it to create representations that are effective for a wide range of tasks. Pre-training acts as a universal feature extractor, providing representations characterised by adaptability and versatility. These serve as a valuable starting point for various tasks, particularly those suffering from low data availability.

\subsection{Large Uni-Modal Models}
\label{sec:bg_unimodal}
Recent research indicates that scaling pre-trained models by increasing both model size and dataset size enhances capacity and generalisation to downstream tasks. While scaling typically retains similar architectures and pre-training tasks, larger models exhibit distinct behaviours. For instance, the 330M parameter BERT \cite{devlin2018bert} and the 1.5B parameter GPT-2 \cite{radford2019language} demonstrate different capabilities, with larger models exhibiting surprising abilities \cite{wei2022emergent}. GPT-3 \cite{brown2020language}, for example, excels at few-shot tasks via in-context learning, a milestone beyond the capabilities of GPT-2. These models employ the same pre-training strategies outlined in Sec. \ref{sec:bg_pretrain}, but scale up in model size, dataset size, and training strategies.

This section focusses on large models typically pre-trained on single-modality datasets. The transformer architecture has revolutionised pre-training in NLP, enabling efficient handling of sequential data, including text, audio, video, and images, outperforming earlier LSTM-based models. In CV, models such as Vision Transformer (ViT) \cite{dosovitskiy2020image}, MAE \cite{he2022masked}, and DINOv1/v2 are trained on extensive datasets such as ImageNet21K \cite{deng2009imagenet} and JFT \cite{sun2017revisiting}. In NLP, large models trained solely on language datasets include BERT \cite{devlin2018bert}, GPT-1/2/3 \cite{brown2020language}, and LLaMA \cite{touvron2023llama}, which are applied to various downstream tasks in both CV and NLP. This paper discusses only the models used by the methods it covers; see \cite{zhao2023survey} for a comprehensive and updated list. 

\subsection{Large Multi-Modal Models}
\label{sec:bg_multimodal}
For a long time, machine learning models operated within uni-modal tasks.
To fully harness their potential, models must integrate multiple modalities and generalise beyond predefined objectives. Recently, many multi-modal systems have been developed to achieve this. CLIP \cite{radford2021learning} was the first model capable of generalising to image classification tasks with zero-/few-shot learning. The first large-scale models in this domain were BLIP \cite{li2023blip} and Flamingo \cite{alayrac2022flamingo}, which further scaled the model size and enabled open-ended responses, marking a significant milestone in the multi-modal domain similar to GPT-3 \cite{brown2020language} in NLP. Visual instruction tuning quickly became a prominent training paradigm in the multi-modal domain, alongside the use of parameter-efficient fine-tuning (PEFT) \cite{xu2023parameter} techniques to adapt these pre-trained (or FMs) for downstream tasks. 
Another notable direction of research is SAM \cite{kirillov2023segment}, pre-trained on segmentation tasks, which demonstrated impressive performance due to its class-agnostic nature and has been applied across various domains, from medical imaging \cite{wu2023medical} to 3D point clouds \cite{chen2024ll3da}.

\subsection{Downstream Task Adaptation}
\label{sec:bg_adaptation}
Downstream task adaptation is achieved through transfer learning \cite{perkins1999transfer} via two main approaches: full fine-tuning, which updates all model parameters, and feature extraction, which adds a task-specific head. Both require significant computational resources and large datasets. With the rise of large models containing billions of parameters, fully fine-tuning has become impractical. To address this, PEFT techniques have been developed, introducing only a few new parameters to effectively adapt the models. For example, prompt-tuning \cite{lester2021power} learns a small set of vectors as soft prompts before input text, while low-rank adaptation (LoRA) \cite{hu2021lora} reduces the number of new weights by learning low-rank matrices. Combined with quantisation methods such as QLoRA \cite{dettmers2024qlora}, these techniques further decrease memory requirements compared to standard half-precision weights.
These strategies enable efficient adaptation of large pre-trained models to a wide range of downstream tasks.

%% file: _sections/3_3d_foundation.tex
\section{Building 3D Foundational Models}
\label{sec:building_3dfms} 

\input{_tables/3d_fms}

Following the definition by \cite{bommasani2021opportunities}, this section provides an overview of initial efforts toward building 3DFMs. 
These approaches employ diverse pre-training strategies and modalities such as RGB images, text, audio, and point clouds to construct robust representations for 3D data. 
Previous work on 3DFMs includes ConvNet-based methods such as PointContrast \cite{xie2020pointcontrast}, OcCo \cite{wang2021unsupervised, hou2021exploring} and DepthContrast \cite{Zhang_2021_ICCV}, alongside transformer-based approaches such as Point-BERT \cite{Yu_2022_CVPR}, Point-MAE \cite{zhang2022point}, Voxel-MAE \cite{hess2022masked}, and SimIP \cite{li2022simipu}. However, these methods fall outside of the scope of this survey as they do not utilise 2DFMs for building 3DFMs and will not be discussed further.

\begin{figure}[!htb]
    \centering
    \includegraphics[width=\linewidth]{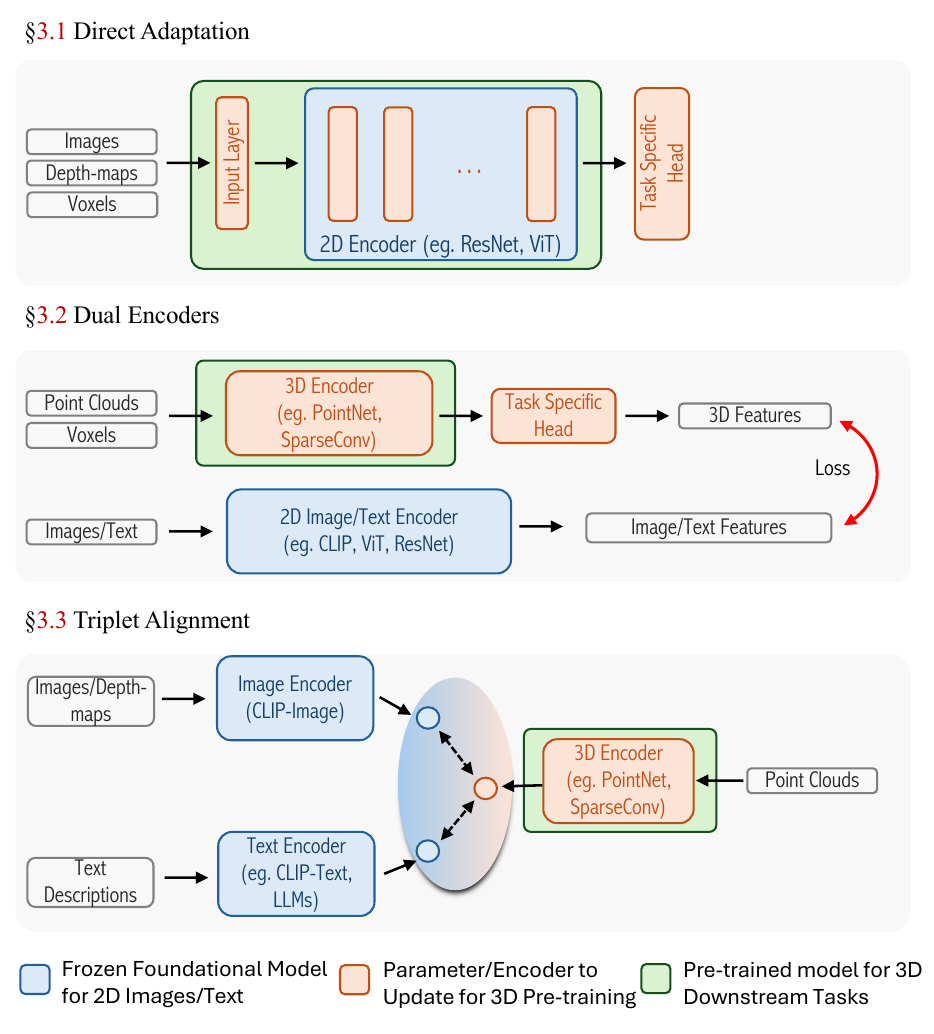}
    \caption{Methods of building 3D models for representation learning by leveraging 2DFMs, categorised by how FMs are used.}
    \label{fig:3fms_groups}
\end{figure}

Given the rapid development of methods aimed at building 3DFMs, it can be challenging to grasp the full scope at once. To facilitate comprehension, we categorise these methods into three groups based on their use of 2DFMs. The first category includes methods that build on top of 2DFMs as a base model. The second group consists of dual encoders, where one encoder is pre-trained 2DFMs and the other encodes 3D data. Lastly, we group methods that build on triplet alignment 
to learn point cloud representations through alignment between text, image, and point cloud encoders. Fig. \ref{fig:3fms_groups} provides a high-level architecture overview and Tab. \ref{tab:3dfms} summarises these methods. Each of these categories is further detailed in the following subsections.

\subsection{Direct Adaptation}
\label{sec:3dfm_direct_adaptation}
Both 2D images and 3D point clouds serve as intuitive visual representations of the physical worlds. Despite significant differences in their underlying representations, they convey similar fundamental visual concepts, and human vision can seamlessly comprehend both forms. Building on this premise, this section explores methods that leverage 2DFMs—such as ConvNets \cite{he2016deep}, ViTs \cite{dosovitskiy2020image} pre-trained on ImageNet-1K/21K \cite{deng2009imagenet, dosovitskiy2020image}, and CLIP \cite{radford2021learning}—to represent point clouds with minimal adjustments, using the same 2DFMs as a base model and without using an entirely new point backbone.

\tbf{Image2Point} \cite{xu2021image2point} demonstrates how commonly used image-pre-trained models, such as 2D ConvNet and ViT, can be effectively adapted for various 3D tasks. Specifically, pre-trained 2D ConvNet and ViT can be transformed into projection-based, voxel-based, and transformer-based point-cloud models by either copying or inflating weights. The approach focusses primarily on 3D ConvNet, which utilises voxel input by inflating 2D kernels into 3D ones derived from 2DFMs. To these transformed models, linear input and output layers are added. During fine-tuning on target point cloud datasets, only the input/output layers and batch normalisation layers are adjusted while the pre-trained model weights remain unchanged. It uses ResNet50 \cite{he2016deep} trained on ImageNet-1K \cite{deng2009imagenet} as the base model; this approach outperforms previous techniques of that time.

Similarly \tbf{Pix4Point} \cite{qian2022improving} learns 3D representations on top of 2DFMs by introducing a point tokenizer at the input and a task-specific head or decoder at the output. Instead of inflating weights as in Image2Point, this method focusses on transformer architecture, which can process various input types through tokenization. The 3D model is initialised from a standard pre-trained transformer and incorporates a relative progressive tokenizer that gradually converts point clouds into tokens using graph convolutions. The model employs a ViT \cite{dosovitskiy2020image} trained in a self-supervised manner as the base model to learn the point cloud representation.

In contrast to previous approaches that rely solely on point clouds as input, \tbf{PCExpert} \cite{kang2023point} leverages images to guide point cloud learning. Inspired by the mixture-of-modality-expert \cite{bao2021vlmo}, which encodes different modalities using modality-specific experts, PCExpert employs a pre-trained ViT to encode both point clouds and images. This is achieved by separating the feed forward network (FFN) and layer normalisation within each transformer \cite{vaswani2017attention} layer for each modality. Additionally, a projection head is added to the output for point cloud processing. The model incorporates a shared multi-head self-attention mechanism to ensure knowledge transfer between the image and point cloud modalities, while the separate FFNs capture the unique features of each modality. During pre-training, the ViT parameters are frozen, and only the parameters related to point clouds and the projection head are updated.

\tbf{PointCLIP} \cite{zhang2022pointclip} leverages the CLIP \cite{radford2021learning} framework for point cloud understanding. It bridges the modality gap by projecting point clouds onto predefined image planes to generate multi-view depth maps, allowing for feature extraction with CLIP’s pre-trained visual encoder, and the text encoder is used for zero-shot classification without dedicated 3D training; however, the method lags behind state-of-the-art methods. 
Additionally, a learnable inter-view adapter is introduced to consolidate features from multiple views, and 
fine-tuned while keeping the CLIP encoders frozen, leading to competitive performance with state-of-the-art methods. \tbf{PointCLIPV2} \cite{zhang2022pointclipv2} improves upon PointCLIP by integrating the strengths of CLIP and GPT-3 \cite{brown2020language} for unified 3D open-world understanding. Enhances zero- and few-shot tasks, including 3D classification, part segmentation, and object detection, without requiring 3D training data.

\tbf{DiffCLIP} \cite{shen2023diffclip} adopts a novel approach for learning point cloud representations using the CLIP \cite{dosovitskiy2020image} model. Since CLIP was originally trained on image-text pairs, there is a significant domain gap between point clouds and images. To address this, DiffCLIP leverages a diffusion model integrated with ControlNet \cite{zhang2023adding}. Given a point cloud as input, on the image side of CLIP, it generates multi-view depth maps. These depth maps are then transformed through a style transfer process, guided by stable diffusion \cite{rombach2022high} and ControlNet, into photo-realistic 2D RGB images, which are fed into the frozen image encoder of CLIP. On the text side, for zero-shot tasks, hand-crafted prompts are applied to both the stable diffusion module and the frozen text encoder. For few-shot tasks, a style-prompt generation module is introduced, generating prompts alongside class labels that are fed into the frozen text encoder of CLIP. Both the frozen image and text encoders are used to produce feature representations of images and text, which are subsequently processed through a multi-modal fusion block. This block computes cosine similarity between the modalities and includes an additional ConvNet block to fuse logits from each view for the final output.

\subsection{Dual Encoders}
\label{sec:3dfm_dual_encoders}

This section highlights methods that utilise 2DFMs for 3D understanding by adding a separate point cloud backbone in parallel to the 2D backbone. These methods are typically trained using contrastive learning, knowledge distillation from the 2D model, and later employ the point cloud backbone for feature extraction in downstream tasks.

The pixel-to-point knowledge transfer (\tbf{PPKT}) \cite{liu2021learning} method leverages 2D information by mapping pixel-level and point-level features into a shared embedding space through a differentiable back-projection function. However, the output of the 2D backbone lacks pixel-level resolution. To address this, an upsampling layer, a modified version of the projection layer used in contrastive learning frameworks \cite{Hadsell2006dimensionality}, is introduced. This upsampling layer restores the spatial resolution of the image feature map to its original size. Simultaneously, the point cloud is processed by a 3D backbone to extract 3D features. During pre-training, the 2D model is frozen, while only the 3D backbone is trained using point-pixel noise contrastive estimation (NCE) \cite{gutmann2010noise} loss.

Similar to PPKT, \tbf{CrossPoint} \cite{afham2022crosspoint} adopts a dual-branch architecture--one branch processes point clouds, while the other processes images. The point cloud branch establishes intra-modal correspondence by ensuring robustness to point cloud augmentations, meaning it can handle variations in the data effectively. Meanwhile, the image branch creates cross-modal correspondence by applying a contrastive loss between the rendered 2D image features and the point cloud prototype features. CrossPoint jointly trains the model by combining the learning objectives of both branches. For downstream tasks, only the point cloud backbone is used, and the image branch is discarded.

\tbf{CLIP2Point} \cite{huang2023clip2point} focusses on aligning depth maps rendered from 3D point clouds with CLIP visual features. This is done using a self-supervised pre-training scheme that incorporates both intra-modality and cross-modality contrastive learning to align depth features with CLIP's visual features. During the process, a pair of rendered depth maps is constructed by randomly selecting camera views for each 3D point cloud and modifying their view distances. The method applies a combination of NT-Xent \cite {chen2020simple} and InfoNCE loss \cite{oord2018representation} to pairs of depth features from the depth encoder and between the depth and image features. The image encoder is kept frozen during training to ensure that depth features align with the CLIP visual encoder. With the aligned depth encoder, the point cloud features are then compared with the CLIP text features for downstream tasks \footnote{This shares some similarity with the triplet alignment discussed in Section 3.3, but since the text encoder is only used for the downstream task, we have included it here.}.

Yao et al. \cite{yao20223d} proposed \tbf{PointCLIPKD}, a point cloud learning method using CLIP. Like CLIP2Point, it has both point cloud and image backbones. However, unlike CLIP2Point, it adds a ClipCap \cite{mokady2021clipcap} module on the image side and directly takes point clouds instead of depth maps on the point cloud side. The point cloud backbone receives two inputs: point cloud tokens and concept query sets. These concept tokens are extracted layer by layer using cross-attention between the image and point cloud branches. Knowledge is transferred by aligning the concept tokens with the image embedding through a distillation loss \cite{hinton2015distilling}.

\tbf{Bridge3D} \cite{chen2023bridging} bridges 2D and 3D vision by pre-training a 3D model using features, semantic masks, and captions extracted from multiple 2D FMs in a self-supervised manner. Semantic masks from the 2D models guide the masking and reconstruction process for MAE \cite{he2022masked}, focussing attention on important foreground objects. At the scene level, an image captioning FMs, such as Tag2Text \cite{huang2023tag2text}, bridges the gap between point clouds and text for scene-level knowledge distillation. Additionally, object-level knowledge distillation leverages precise object masks and semantic text from 2D FMs.

Zhang et al. \cite{zhang2023learning} proposed \tbf{I2P-MAE}, a method for transferring knowledge from images to points for self-supervised pre-training of point clouds in MAE \cite{he2022masked} style. This method projects point clouds into multi-view depth maps to obtain 2D semantics of 3D shapes. Instead of randomly masking point clouds for the MAE input, a saliency map from 2DFMs guides the point cloud masking, focussing on key visible structures to comprehend the global 3D shape. During training, 2D semantics are reconstructed from visible point tokens, ensuring that both 3D spatial patterns and high-level semantic information from 2D models are captured, enhancing 3D representation.

\subsection{Triplet Alignment}
\label{sec:3dfm_triplet_aligment}
This section reviews methods that focus on aligning point clouds, images, and text together. Most of these approaches are built on top of CLIP, incorporating a new encoder for point clouds and aligning it with the frozen encoders of CLIP. Subsequently, the point cloud encoder is utilised to extract point cloud features for downstream tasks.

\tbf{ULIP} \cite{xue2023ulip} seeks to unify representations by aligning all three modalities--image, text, and point cloud. The inputs to ULIP consist of objects represented as triplets of images, text, and point clouds. The image and text features are extracted from the pre-aligned frozen vision and language encoders of CLIP \cite{radford2021learning}, while the 3D features are extracted from a randomly initialised point cloud encoder. Contrastive losses \cite{Hadsell2006dimensionality} are applied to align the 3D feature of an object with its image and text features during pre-training. The point cloud encoder is then used to extract point cloud features for downstream tasks. ULIP shows strong performance as an end-to-end method for point cloud pre-training. However, the textual descriptions in ULIP are relatively shallow, lacking the fine-grained information and details necessary for a comprehensive understanding. To address this limitation, \tbf{ULIP-2} \cite{xue2023ulip2} introduces the use of large vision-language models (LVLMs), such as BLIP-2 \cite{li2023blip}, which improve scalability and reduce data requirements. 

CLIP Goes 3D  (\tbf{CG3D}) \cite{hegde2023clip} was proposed around the same time as ULIP and shares similar goals and training strategies. 
However, CG3D observes a distribution shift in 2D rendered images within the CLIP 3D setting. To address this, CG3D introduces a learnable prompt and incorporates additional training parameters in the image input space to align CLIP with the 3D pre-training dataset. The effectiveness of the pre-trained 3D encoder is demonstrated in tasks such as open scene understanding and retrieval.

Similarly, \tbf{CLIP2Scene} \cite{chen2023clip2scene} uses CLIP's image and text encoders to train a 3D encoder in cross-modal learning tasks. It regularises a 3D network by incorporating CLIP's semantic and visual information. CLIP2Scene introduces semantic consistency regularisation (SCR) and spatial-temporal consistency regularisation (STCR) modules. For SCR, CLIP’s text semantics are used to select positive and negative point samples for conflict-free contrastive learning. For STCR, CLIP’s image pixel features impose a soft consistency constraint on temporally coherent point features. The trained 3D encoder is applied to indoor and outdoor datasets for downstream semantic segmentation tasks.

\tbf{Text4Point} \cite{huang2024joint} focusses on integrating the language modality into 3D vision models, bridging the gap between point clouds and text by leveraging 2D images to learn point cloud representations. The model establishes correspondences between images and point clouds. Using CLIP’s well-aligned image and text features, point cloud features are implicitly aligned with text embeddings. To enable the 3D model to leverage text information from CLIP’s pre-trained text encoder, the embedding spaces of the CLIP 2D encoder and the 3D encoder are aligned through a pixel-voxel contrastive loss \cite{Hadsell2006dimensionality}. If needed, the 2D and 3D decoder features are further aligned for downstream tasks like semantic segmentation, instance segmentation, and object detection.

Drawing inspiration from CLIP's applications in open-world understanding, \tbf{CLIP\tss{2}} \cite{zeng2023clip2} pre-trains a 3D encoder to acquire a general representation of 3D data. Instead of projecting 3D point clouds into images, CLIP\tss{2} directly aligns the 3D space with the raw text to learn a 3D representation in an open-world setting. This model collects one million triplets of point clouds, images, and text by employing 2DFMs. The feature space alignment of the three modalities is optimised jointly, encompassing the semantic-label text-3D correlation and the instance-level image-3D correlation. The model demonstrates promising results for zero-shot recognition on indoor and outdoor datasets, especially for long-tail categories.

\tbf{OpenShape} \cite{liu2023openshape}, similar to ULIP \cite{xue2023ulip}, focusses on improving the understanding of open-world 3D shapes. To achieve this, OpenShape scales up the 3D dataset by integrating four public 3D shape datasets, covering a wide range of categories. In addition, it improves the quality of the text through filtering, captioning, and image retrieval strategies to automatically refine and improve text descriptions. The 3D backbone is scaled to suit large 3D datasets, and data resampling is handled using hard negative mining, improving the model's discrimination ability. OpenShape is tested on a zero-shot classification task, demonstrating its efficacy in open-world recognition.

Although most methods discussed so far have focused on pre-training models for tasks such as classification, detection, and segmentation, \tbf{Multi-CLIP} \cite{delitzas2023multi} facilitates learning language-grounded and transferable 3D scene representations. Multi-CLIP aligns 3D scene features with 2D multi-view images and text embeddings using a contrastive objective in the CLIP space. The model is evaluated in challenging downstream tasks, such as 3D visual question answering (3D-VQA) and 3D situated question answering (3D-SQA).

\tbf{Uni3D} \cite{zhou2023uni3d} addresses the limitations of existing 3D pre-training methods, offering a unified and scalable framework for large-scale 3D representation learning. By leveraging a 2D ViT as the 3D encoder and initialising it with the best 2D priors, Uni3D undergoes end-to-end pre-training to align 3D point cloud features with image-text aligned features. With a billion parameters, a million 3D shapes, and ten million images paired with 70 million texts, Uni3D explores scalability in 3D representation learning. It utilises abundant 2D pre-trained models like Eva \cite{fang2023eva} and DINO \cite{caron2021emerging}, and image-text aligned models such as CLIP \cite{radford2021learning} and scaling law \cite{cherti2023reproducible}, demonstrating continuous performance improvements as the model scales.

Applying 2D alignment strategies to 3D data presents challenges such as information degradation, insufficient synergy between 3D, image, and text features, and under-utilisation of fine-grained details. \tbf{JM3D} \cite{ji2023jm3d} addresses these issues, information degradation by integrating point cloud, text, and image information using a structured multi-modal organiser which generates a continuous sequence of multi-view rendered images and establishes a hierarchical text tree. The inadequate synergy is addressed by seamlessly integrating textual and visual modalities, which leads to a unified representation through joint multi-modal alignment. This approach is extended in \tbf{JM3D-LLM} \cite{ji2023jm3d}, embedding 3D representations into LLMs for tasks such as image-3D retrieval and zero-shot 3D classification, providing more granular information.

%% file: _tables/3d_fms.tex
\begin{table*}[!ht]
    \centering
    \tiny
    \caption{Summary of methods that aim to build a 3D foundational (or representation) model by utilising 2D foundational models.}
    \begin{tabular}{llrp{1.5cm}p{2.1cm}p{3cm}p{1cm}p{1cm}p{1.8cm}}
    \toprule
        \multirow{2}{*}{\tbf{Types}} & \multirow{2}{*}{\tbf{Method}} & \multirow{2}{*}{\tbf{Venue}} & \multirow{2}{*}{\tbf{2D Backbone}} & \multirow{2}{*}{\tbf{3D Backbone}} & \multirow{2}{*}{\tbf{3D Pre-train Dataset}} & \multicolumn{2}{c}{{\tbf{Zero-shot Classification Acc}}} & \multirow{1}{*}{\tbf{Part Segmentations}} \\
        \cmidrule(lr){7-8} \multicolumn{1}{c}{} & \multicolumn{1}{c}{} & \multicolumn{1}{c}{} & \multicolumn{1}{c}{} & \multicolumn{1}{c}{} & \multicolumn{1}{c}{} & ModelNet40 & ScanObjectNN & ShapeNetPart (mIOU) \\
    \midrule
        \multirow{6}{*}{Direct Adaptation} & Image2Point \cite{xu2021image2point} & ECCV'22 & ResNet \cite{he2016deep}, ViT \cite{dosovitskiy2020image} & - & ImageNet1K \cite{deng2009imagenet} & - & - & - \\ 
        ~ & Pix4Point \cite{qian2022improving} & 3DV'24 & ViT \cite{dosovitskiy2020image} & - & ImageNet1K \cite{deng2009imagenet} & - & - & 86.8 \\ 
        ~ & PointCLIP \cite{zhang2022pointclip} & CVPR'22 & CLIP \cite{radford2021learning} & - & CLIP-dataset  \cite{radford2021learning} & 20.18 & 14.12 & 31 \\ 
        ~ & PointCLIPV2 \cite{zhang2022pointclipv2} & ICCV'23 & CLIP \cite{radford2021learning} & - & CLIP-dataset \cite{radford2021learning} & 29.71 & 18.18 & 49.5 \\ 
        ~ & PCExpert \cite{kang2023point} & Arxiv'23 & ViT \cite{dosovitskiy2020image} & - & ShapeNet \cite{wu_3d_2015} & - & - & - \\ 
        ~ & DiffCLIP \cite{shen2023diffclip} & Arxiv'23 & CLIP \cite{radford2021learning}, ControlNet \cite{zhang2023adding}, StableDiffusion \cite{rombach2022high} & - & CLIP-dataset \cite{radford2021learning}, LAION \cite{schuhmann2022laion} & - & - & - \\
    \midrule
        \multirow{7}{*}{Dual Encoders} & PPKT \cite{liu2021learning} & Arxiv'21 & ResNet \cite{he2016deep}  & SparseUNet \cite{choy20194d} & ScanNet \cite{dai_scannet_2017} & - & - & - \\ 
        ~ & CrossPoint \cite{afham2022crosspoint} & CVPR'22 & ResNet \cite{he2016deep} & PointNet \cite{qi2017pointnet}, DGCNN \cite{wang2019dynamic} & ShapeNet \cite{wu_3d_2015} & - & - & 85.5 \\ 
        ~ & CLIP2Point \cite{huang2023clip2point} & ICCV'23 & CLIP \cite{radford2021learning} & Transformer \cite{vaswani2017attention} & ShapeNet \cite{wu_3d_2015} & 33 & 23.32 & - \\ 
        ~ & PointCLIPKD \cite{yao20223d} & Arxiv'22 & CLIP \cite{radford2021learning}, ClipCap \cite{mokady2021clipcap} & Transformer \cite{vaswani2017attention} & ScanNetv2 \cite{dai_scannet_2017} & - & - & 86.1 \\ 
        ~ & I2P-MAE  \cite{zhang2023learning} & CVPR'23 & CLIP \cite{radford2021learning} & MAE (Transformer) \cite{he2022masked} & ShapeNet \cite{wu_3d_2015} & - & - & 86.76 \\ 
        ~ & Bridge3D \cite{chen2023bridging} & NeurIPS'23 & CLIP \cite{radford2021learning}, DINOV2 \cite{oquab2023dinov2}, Tag2Text \cite{huang2023tag2text} & MAE (Transformer) \cite{he2022masked} & ScanNet \cite{dai_scannet_2017}, SUN RGB-D \cite{song2015sunrgbd} & - & - & - \\
    \midrule
        \multirow{18}{*}{Triplet Alignment} & ULIP \cite{xue2023ulip} & CVPR'23 & CLIP \cite{radford2021learning} & PointNet \cite{qi2017pointnet}  & Triplet Collection form ShapeNet \cite{wu_3d_2015} & 60.4 & 48.5 & - \\ 
        ~ & ULIP2 \cite{xue2023ulip2} & Arxiv'23 & CLIP \cite{radford2021learning}, BLIP2 \cite{li2023blip} & PointBERT \cite{yu2022point}, PointNeXt \cite{qian2022pointnext} & Triplet Collection form ShapeNet \cite{wu_3d_2015}, Objectverse \cite{deitke2023objaverse} & 69.7 & ~ & - \\ 
        ~ & CG3D \cite{hegde2023clip}& CVPR'23 & CLIP \cite{radford2021learning} & PointTransformer \cite{zhao2021point}, PointMLP \cite{ma2022rethinking} & ShapeNet \cite{wu_3d_2015} & 50.6 & 25.6 & - \\ 
        ~ & CLIP2Scene \cite{chen2023clip2scene} & CVPR'23 & CLIP \cite{radford2021learning} & SPVConv \cite{tang2020searching}, SparseUNet \cite{choy20194d} & NuScenes \cite{caesar_nuscenes_2020} & - & - & - \\ 
        ~ & Text4Point \cite{huang2023joint} & PR'23 & CLIP \cite{radford2021learning} & SparseUNet \cite{choy20194d} & ScanNet \cite{dai_scannet_2017} & - & - & - \\ 
        ~ & CLIP\tss{2} \cite{zeng2023clip2} & CVPR'23 & CLIP \cite{radford2021learning} & PointNet \cite{qi2017pointnet}, PointNet++ \cite{qi2017pointnet++} & Triplet Collection \cite{zeng2023clip2} & 37.8 & 39.1 & - \\ 
        ~ & OpenShape \cite{liu2023openshape} & Arxiv'23 & CLIP \cite{radford2021learning}, BLIP \cite{li2023blip} & PointBERT \cite{yu2022point}, PointNeXt \cite{qian2022pointnext} & Combination of ShapeNet \cite{wu_3d_2015}, 3D-Future \cite{fu20213d}, ABO \cite{collins2022abo} , Objectverse \cite{deitke2023objaverse} & 85.3 & 47.2 & - \\ 
        ~ & Multi-CLIP \cite{delitzas2023multi} & Arxiv'23 & CLIP \cite{radford2021learning} & PointNet++ \cite{qi2017pointnet++} & ScanRefer \cite{chen2020scanrefer}, ScanNet \cite{dai_scannet_2017} & - & - & - \\ 
        ~ & Uni3D \cite{zhou2023uni3d} & Arxiv'23 & CLIP \cite{radford2021learning}, EVA \cite{fang2023eva}, DINO \cite{caron2021emerging} & ViT \cite{dosovitskiy2020image} & Combination of ShapeNet \cite{wu_3d_2015}, 3D-Future \cite{fu20213d}, ABO \cite{collins2022abo} , Objectverse \cite{deitke2023objaverse} & 87.3 & 63.9 & 78.2 \\ 
        ~ & JM3D \cite{ji2023jm3d} & Arxiv'23 & CLIP \cite{radford2021learning} & PointNet \cite{qi2017pointnet}, PointBERT \cite{yu2022point}, PointMLP \cite{ma2022rethinking} & ShapeNet \cite{wu_3d_2015} & 55.5 & 47.5 & 83.6 \\
    \bottomrule
    \end{tabular}
    \label{tab:3dfms}
\end{table*}

%% file: _sections/4_2d_vlms_for_3d.tex
\section{Adapting 2DFMs for 3D}
\label{sec:perception_tasks}

For the categories in this area, most of the methods overlap with those described in Section \ref{sec:building_3dfms}. 
To minimise the repetition, we focus on the methods
that use 2DFM for a specific task even with pre-training, i.e. even if it is pre-trained, it can only be used for that specific task.

\subsection{2DFMs for 3D Classification}
\label{sec:classification}

\input{_tables/classification}
This section reviews methods for point cloud classification that leverages 2DFMs. 

Similarly, \tbf{CALIP} \cite{guo2023calip} adapts CLIP models without any learnable parameters or training. It uses a parameter-free cross-modal attention mechanism to align spatial visual and textual features and introduces non-parametric and bidirectional modifications, similar to PointCLIP. In the non-parametric approach, linear layers are omitted, relying solely on well-aligned CLIP features to compute attention maps. In the bidirectional approach, both visual and textual features are updated simultaneously, with visual features guided by category semantics from text and textual features made visual-aware and image-conditional.

\tbf{P2P} \cite{wang2022p2p} offers a different approach, introducing a prompting technique inspired by VPT \cite{jia2022visual} to transfer 2D pre-trained knowledge into the 3D domain. It transforms point clouds into colourful images via geometry-preserved projection and geometry-aware colouring. These images are processed by a frozen pre-trained image model to extract features, which are then used by downstream task-specific heads. This method promotes bidirectional knowledge flow between points and pixels, retaining geometric information through projection and infusing colour information back into the colourless point clouds via the interaction between the geometry-aware colouring module and the pre-trained image model.


In contrast to the text-prompting methods, \tbf{MvNet} \cite{peng2023multi} uses multi-view projected features to prompt a pre-trained vision model. The point cloud is encoded into multiple views, and a multi-view prompt vision fusion module interchanges and merges information across views through an attention mechanism. This produces prompts that leverage prior knowledge from a 2DFM for 3D few-shot learning.

Adapting 2D pre-trained models for point cloud recognition through multi-view projection is effective, but they can introduce challenges such as incomplete geometric information and rendering artifacts. To address this, \tbf{InvJoint} \cite{yi2023invariant} proposes leveraging both 2D and 3D domains simultaneously. By analysing the confusion matrices of 2D and 3D models, the authors demonstrate that each model is confused by different classes, limiting the effectiveness of late fusion. Their method selects joint hard samples using a Gaussian mixture model (GMM) \cite{reynolds2009gaussian}, followed by a joint learning module that captures collaborative representations across domains through an invariant feature selector.

Table \ref{tab:classification} shows the overview of these methods and their performance on ModelNet40 and ScanObjectNN in different shots.


\subsection{2DFMs for 3D Segmentation}
\label{sec:segmentation}
In this section, we explore 2DFMs such as CLIP and SAM for point cloud segmentation. As most FMs-focused works revolve around open-vocabulary or open-world 3D segmentation, we summarise methods specifically targeting these areas. Additionally, we briefly discuss works that utilise 2DFMs for part segmentation.

\subsubsection{3D Part Segmentation} 
\input{_tables/part_segmentation}
The introduction of VL models such as CLIP and segmentation models like SAM has unleashed significant advancements in 2D CV tasks. Leveraging this progress, recent work has extended the capabilities of these 2DFMs to tackle 3D part segmentation challenges. 
\tbf{PointCLIP V2} \cite{zhang2022pointclipv2} stands as an early example, achieving zero-shot part point cloud segmentation solely through 2D vision and text backbones. 
\tbf{PartSLIP} \cite{liu2023partslip} further builds on this foundation by utilising GLIP, an enhanced version of CLIP, to localise 2D objects based on free-form text and employ a 3D voting and grouping module, effectively transitioning multi-view 2D bounding boxes into 3D semantic and instance segmentation.
\tbf{ZeroPS} \cite{xue2023zerops} takes a similar approach, utilising SAM and GLIP alongside multi-view perspectives to achieve zero-shot 3D part segmentation. It expands 3D groups from local to global levels, introduces a merging algorithm for part-level groups, integrates a two-dimensional checking mechanism to identify the best matching 3D part for each 2D box, and incorporates a class non-highest vote penalty function to refine the vote matrix.
In the case of \tbf{PartDistill} \cite{umam2023partdistill}, a departure from direct use of the 2D backbone is observed. Instead, it adopts a knowledge distillation mechanism. This involves a teacher network employing a 2DFMs for 2D predictions and a student network learning from these predictions while extracting geometric features from multiple 3D shapes to facilitate 3D part segmentation.

\input{_tables/instance_segmentation}
In addition to 3D part segmentation, researchers have harnessed the capabilities of robust 2DFMs, such as SAM, to segment more intricate scenes. 
One line of approaches focusses on accurately segmenting 2D frames using various scene deconstruction techniques. For instance, \tbf{SAM3D} \cite{yang2023sam3d} utilises multi-view images as input, employing SAM to generate 2D masks which are then mapped to 3D. This method iteratively merges adjacent point clouds with the bidirectional merging approach until obtaining 3D masks for the entire scene. Similarly, \tbf{SAM-Graph} \cite{guo2023sam} over-segments input mesh/point cloud into superpoints, constructing a graph structure based on adjacency. SAM annotates nodes and edges, while node features aggregate multi-view SAM backbone features and edge weights rely on intersection ratios between superpoint masks. A graph neural network refines the graph, executing a graph cut for instance segmentation. \tbf{SAI3D} \cite{yin2023sai3d} is another variant that utilises a similar technique for instance segmentation without the graph neural network. 
Another line of approaches is learning high-quality 3D points to prompt SAM using 3D projections, such as \tbf{SAMPro3D} \cite{xu2023sampro3d}. Given a 3D point cloud and multiple posed RGB frames, SAM is applied on RGB frames to segment 3D scenes without requiring training data. 
Furthermore, \tbf{PointSeg} \cite{he2024pointseg} adopts a two-branch prompts learning structure, featuring bidirectional matching-based prompts generation, iterative prompt-refinement, and affinity-aware merging, enhancing FMs' ability to improve 3D segmentation quality. 


\subsubsection{Open-Vocabulary Segmentation}
\begin{figure}[!htb]
  \centering
  \includegraphics[width=\linewidth]{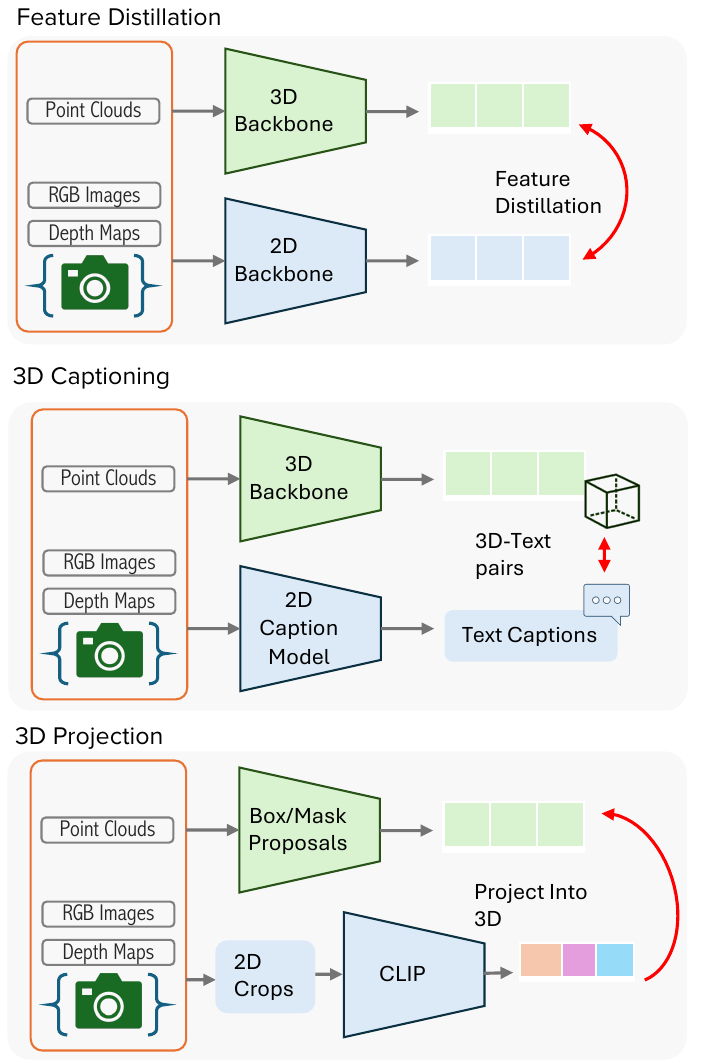}
  \caption{High level overview of open vocabulary methods for point clouds using 2DFMs.}
  \label{fig:general_pipeline}
\end{figure}

\input{_tables/open_vocab}
Open-vocabulary (OV) representation methods can naturally address the settings of open-set and open-world.
This task extends to open-set classification, introducing an additional generalisation requirement to novel vocabulary, visual properties, and domains. The inherent challenge lies in the limited diversity and scale of 3D-text pairs datasets compared to their 2D counterparts. Training on such 3D data may not adequately prepare AI systems for the complexities of the open 3D world. These methods can be divided into three groups: Feature distillation, Captioning based, and 2D projection methods. 

\underline{\tbf{Feature Distillation}} bridges the gap between 2D and 3D representations by aligning language-associated features with 3D representations.
For example, \tbf{OpenScene} \cite{peng2023openscene} uses 2D pixel labels for 3D comprehension, avoiding the need for explicit 3D annotations. It computes dense features for 3D points, embedding them with text and image pixels in the CLIP feature space. By aligning 3D points with pixels from posed images, it trains a 3D network to embed points using CLIP pixel features. This allows open vocabulary queries on 3D points, facilitated by multi-view fusion of pixel features. A sparse 3D ConvNet is trained to extract features from point cloud geometry, reducing differences to aggregated pixel features. The final step combines 2D and 3D features into a hybrid strategy, enabling robust 3D scene understanding queries, capturing concepts like affordances, materials, and functions in tasks such as segmentation, detection, and exploration.

Similarly, \tbf{CLIP-FO3D} \cite{zhang2023clip} eliminates the need for point cloud label, image pixel supervision, or pseudo-caption. Instead, it directly transfers CLIP’s knowledge to 3D models without annotations. CLIP-FO3D addresses the challenge of applying CLIP to 3D vision by extracting pixel-level features, overcoming the limitations of global features and 3D scene complexity. Building on MaskCLIP, it crops images at multiple scales to preserve object semantics and extracts pixel-level features from each RGB view. Using the 3D multi-view projection scheme, these features are mapped onto the point cloud, serving as training targets aligned with the CLIP feature space. Through feature distillation, the 3D model minimises the gap between learnt and target features, enabling CLIP-FO3D to extract open-world 3D scene representations. In particular, it excels in annotation-free OV 3D semantic segmentation, using text embeddings of class names as classification weights.

\underline{\tbf{3D Captioning}} methods focus on building 3D-text pairs and training a 3D instance proposal network alongside a contrastive OV model. This approach aligns the predicted proposals with their corresponding text captions.
\tbf{PLA} \cite{ding2022pla} employs an image-captioning model to generate captions aligned with 3D data, enhancing representation through text embedding while training a 3D network to learn language-aware embeddings from these pseudo-captions. The challenge lies in the complex object compositions in 3D scene-level data, making it difficult to link objects to their corresponding words in the captions. Unlike object-centric images, multi-view images of a 3D scene are related by geometry, enabling the creation of hierarchical point-caption pairs at scene, view, and entity levels. These pairs provide coarse-to-fine supervision, facilitating the learning of visual-semantic representations for OV 3D scene understanding tasks like semantic and instance segmentation. The PLA point-language association paradigm is versatile for such tasks without requiring a task-specific design.

In \tbf{RegionPLC} \cite{yang2024regionplc}, it generates 3D language pairs at the region level by combining various region-level captions from 2DFMs, such as image captioning, dense captioning, and detection models. These captions are mapped to 3D regions to create paired data. The framework includes a region-aware point-discriminative contrastive learning mechanism that prevents confusion between unrelated points in close proximity, enhancing the discriminatory power of point-wise embeddings. It also normalises the learning process across regions of varying sizes, making feature learning more robust. The final model integrates with language models for open-ended 3D reasoning, achieving grounding capabilities without the need for task-specific training data.

\underline{\tbf{3D Projection}} methods obtain the 2D features and/or mask using methods such as CLIP \cite{radford2019language} and SAM \cite{zhang2023segment}, and are back-projected into 3D. 
An example of this, \tbf{SemAbs} \cite{hasemantic}, addresses visual-semantic reasoning in open-world 3D scene understanding tasks using 2D VLMs. Its hypothesis posits that, while open-world visual-semantic reasoning necessitates exposure to internet-scale datasets, 3D spatial and geometric reasoning is tractable even with a limited synthetic dataset and could generalise more effectively if learnt in a semantic-agnostic manner. Instead of learning the complete concept, the model focusses on abstract concepts. To achieve this level of abstraction, it leverages relevancy maps extracted from CLIP, using representations that remain agnostic to semantic labels. An evaluation was conducted on the completion of the semantic scene of the OV and the location of visually obscured objects. Despite training only the 3D network on a restricted synthetic dataset, the model demonstrates the ability to generalise to any novel semantic labels, vocabulary, visual properties, and domains to which the 2DFMs can effectively generalise. 

\tbf{OpenMask3D} \cite{takmaz2023openmask3d} build on top of Mask3D, distinguishes itself from existing 3D OV methods by employing an instance-based feature computation approach instead of a point-based one. Given an RGB-D sequence and the corresponding 3D geometry, OpenMask3D predicts 3D object instance masks and computes a mask-feature representation. Its two-stage pipeline includes a class-agnostic mask proposal head and a mask-feature aggregation module. The aggregation module identifies top-k frames with high instance visibility, extracts CLIP features from the best images using a multi-scale, crop-based approach. By aggregating these features across views, they obtain a feature representation for each 3D instance mask.  This allows for the retrieval of object instance masks based on similarity to any given query, facilitating open-vocabulary 3D instance segmentation. It shows that the zero-shot nature of feature computation enhances the preservation of information about novel and long-tail objects compared to trained counterparts. 

\textbf{OpenIns3D} \cite{huang2023openins3d} is a 3D open-vocabulary framework designed for versatile deployment without relying on well-aligned 2D images. It involves three key steps: Mask, Snap, and Lookup. The mask proposal module generates class-agnostic 3D mask proposals, evaluates them with the mask scoring module, and filters invalid masks. Synthetic scene-level images are then generated using calibrated camera poses, minimising rendering needs. These images are processed by 2D OV models, creating a class lookup table to store detected categories. The Mask2Pixel maps project 3D proposals onto 2D images, allowing category assignment during Lookup. The results of multiple views are combined for the initial classification, with refinements to remove unassigned masks.

\tbf{OVIR-3D} \cite{lu2023ovir} introduces an OV instance retrieval method that ranks 3D instance segments from a point cloud reconstructed using RGB-D video and a language query. By leveraging multi-view fusion and smoothing techniques, it surpasses the limitations of 2D segmentation for more precise 3D object identification. Instead of relying on pixel-level data or additional training, it integrates instance-level information directly into the 3D scene. The method generates 2D object region proposals via a 2D OV detector, followed by data association and filtering to enhance masks and reduce noise. Experiments on video datasets show its efficiency, with a 2D-to-3D fusion module that achieved around 30 fps and near-instant text query inference.

Methods like OpenMask3D and OVIR-3D use pre-trained 2D OV models to generate 2D instance masks, which are projected onto 3D point clouds but often suffer from misalignments. \textbf{Open3DIS} \cite{nguyen2024open3dis} overcomes this by aggregating 2D masks across multiple frames and using 3D-aware feature extraction for precise alignment with text queries. A class-agnostic 3D instance segmenter generates initial 3D proposals, refined by combining 2D masks with superpoints. The refined proposals are integrated with the initial ones, and the Pointwise Feature Extraction module aligns point cloud CLIP features with text embeddings to produce accurate instance masks.

\textbf{Segment3D} \cite{huang2025segment3d} introduces a two-stage training framework for class-agnostic 3D segmentation, eliminating the need for manually labelled 3D data. Unlike earlier methods such as SAM3D \cite{yang2023sam3d}, which rely on complex and error-prone merging processes, Segment3D offers a simplified and direct 3D segmentation pipeline.
In the first stage, the model is pre-trained on partial RGB-D point clouds using 2D segmentation masks automatically generated by SAM. These masks, projected from 2D RGB-D images into the 3D domain, enable the model to learn 3D structures by leveraging large-scale RGB-D datasets without manual annotations. In the second stage, the model is fine-tuned on full 3D scenes to bridge the domain gap between partial and complete 3D point clouds. High-confidence mask predictions from the pre-trained model serve as training signals, improving segmentation accuracy on complete 3D reconstructions and enhancing real-world applicability.

\tbf{Open-YOLO3D} \cite{boudjoghra2024open} presents a more efficient alternative to traditional methods that depend on computationally expensive foundation models such as SAM and CLIP by integrating joint 2D-3D reasoning using 2D bounding-box predictions. An open-vocabulary 2D object detector generates bounding boxes with class labels for each RGB frame associated with the 3D scene, which eliminates the need for heavy 3D proposal generation and provides faster results than CLIP-based methods. A 3D instance segmentation network then generates class-agnostic instance masks for the point clouds. The predicted bounding boxes are used to create a low granularity label map for each RGB frame, assigning class labels to the bounding box areas. Finally, the 3D point cloud is projected onto these LG label maps using intrinsic and extrinsic parameters, enabling efficient class assignment to 3D instances without the complexity of lifting 2D features into 3D.

\subsection{2DFMs for 3D Object Detection}
\label{sec:detection}
This section summarises methods using 2DFMs for 3D object detection, a less explored area compared to semantic segmentation, which provides both detection and segmentation. The emergence of segmentation FMs like SAM, a 2DFM capable of flexible image segmentation, has driven greater progress in 3D segmentation than in object detection.

Inspired by SAM’s application in 3D tasks, \textbf{SAM3D} \cite{zhang2023sam3d} adapts this model for 3D object detection in outdoor scenarios. The approach first projects LiDAR points into colourised bird's eye view (BEV) images using a predefined colour palette, then post-processes these BEV images to optimise them for SAM's requirements. After segmenting these images with SAM, it refines the resulting noisy masks and ultimately predicts 3D bounding boxes with the support of the original LiDAR points.

\textbf{VFMM3D} \cite{ding2024vfmm3d} combines two 2DFMs—SAM for segmentation and DAM for depth estimation—to address challenges in monocular 3D object detection. It generates high-quality pseudo-LiDAR data, enriching monocular images with semantic and depth information without requiring dataset-specific fine-tuning. A sparsification technique reduces the noise in the pseudo-LiDAR data, improving computational efficiency. Compatible with various LiDAR-based 3D detectors, VFMM3D enhances 3D spatial information by integrating depth and semantic cues.

Traditional approaches to OV 3D detection face substantial challenges due to the scarcity of comprehensive 3D-text datasets, limiting the integration of OV capabilities typically derived from vision-text pairs. To overcome these constraints, \textbf{FM-OV3D} \cite{zhang2024fm} introduces a cross-modal knowledge integration framework, using FM such as SAM, CLIP, and GPT-3 \cite{brown2020language}. This methodology focusses on improving object localisation and recognition in 3D models. For object localisation, FM-OV3D utilises SAM’s segmentation capabilities to produce 2D bounding boxes, effectively strengthening 3D object positioning. In terms of recognition, it combines 3D detector point cloud features with textual features generated by GPT-3 and visual features derived from stable diffusion, unified within CLIP’s shared feature space. This alignment of features across modalities enables OV detection by bridging these diverse information domains. Furthermore, the adaptable structure of FM-OV3D supports the application to any OV training set by dynamically generating language and visual prompts for specified classes, expanding its versatility across various tasks.

%% file: _tables/classification.tex
\begin{table*}[!ht]
    \label{tab:classification}
    \caption{Summary of methods for point cloud classification on ModelNet40 and ScanObjectNN in different shots.}
    \centering
    \begin{tabular}{lrrrrrrrrrr}
    \toprule
        \multirow{2}{*}{\tbf{Method}} & \multicolumn{5}{c}{\multirow{1}{*}{\tbf{ModelNet40}}} & \multicolumn{5}{c}{\multirow{1}{*}{\tbf{ScanObjectNN}}} \\
         \cmidrule(lr){2-6} \cmidrule(lr){7-11}
        & 0-shot & 4-shot & 8-shot & 16-shot & All & 0-shot & 4-shot & 8-shot & 16-shot & All \\
    \midrule
        PointCLIP \cite{zhang2022pointclip} & 20.18 & 77.07 & 81.35 & 87.2 & - & 15.38 & 46.14 & 50 & 55.5 & - \\
        P2P \cite{wang2022p2p} & - & - & - & - & 94.00 & - & - & - & - & 89.3 \\
        CALIP \cite{guo2023calip} & 21.47 & - & - & - & - & 16.9 & - & - & - & - \\
        MvNet \cite{peng2023multi} & - & 84.88 & 86.1 & 91.16 & - & - & 59.47 & 68.77 & 75.36 & - \\
        InvJoint \cite{yi2023invariant} & - & 78.95 & 83.61 & 88.97 & - & - & 48.1 & 53.36 & 57.02 \\
    \bottomrule
    \end{tabular}
\end{table*}

%% file: _tables/part_segmentation.tex
\begin{table*}[htbp]
  \centering
  \caption{3D Zero-Shot part segmentation results on the PartNetE \cite{Mo_2019_CVPR} dataset. Object category mIoU (\%) results are shown, with `Overall' denoting results across all 45 categories}
  \resizebox{\linewidth}{!}{
    \begin{tabular}{lrr|cccccccccccccccc}
    \toprule
    \textbf{Methods} & \textbf{Venue} & \textbf{Overall (45)} & \textbf{Chair} & \textbf{Clock} & \textbf{Dishwasher} & \textbf{Door}  & \textbf{Knife} & \textbf{Refrigerator} & \textbf{Table} & \textbf{Box}   & \textbf{Bucket} & \textbf{Lighter} & \textbf{Oven}  & \textbf{Pen}   & \textbf{Safe}  & \textbf{Stapler} & \textbf{Suitcase} & \textbf{Toaster} \\
    \midrule
    PointClip V2 \cite{zhang2022pointclipv2} & ICCV'23 & 16.1  & 30.8  & 0.9   & 6.9   & 20.7  & 26.7  & 9.3   & 6.1   & 32.5  & 3.5   & 13.0  & 7.8   & 16.9  & 3.6   & 20.0  & 5.6   & 0.3  \\
    PartSlip \cite{liu2023partslip} & CVPR'23 & 34.4  & 77.1  & 17.1  & 30.5  & 35.7  & 31.2  & 35.7  & 46.0  & 60.5  & 22.1  & 34.3  & 34.1  & 5.6   & 14.8  & 26.4  & 50.8  & 10.7  \\
    ZeroPS \cite{xue2023zerops} & arXiv'23 & 39.3  & 73.1  & 29.7  & 47.2  & 27.5  & 49.6  & 47.7  & 41.6  & 53.9  & 74.8  & 47.2  & 27.2  & 18.5  & 20.0  & 38.7  & 62.9  & 17.8  \\
    PartDistill \cite{umam2023partdistill} & CVPR'24 & 39.9  & 74.1  & 23.6  & 18.6  & 41.1  & 59.2  & 25.2  & 50.2  & 69.7  & 16.8  & 37.3  & 34.2  & 15.7  & 18.2  & 65.1  & 43.2  & 11.4 \\
    \bottomrule
    \end{tabular}
    }
  \label{tab:part_segmentation}
\end{table*} 

%% file: _tables/instance_segmentation.tex
\begin{table}[tb]
  \caption{Performance of 3D instance segmentation on ScanNet \cite{dai2017scannet}, ScanNet++ \cite{yeshwanth2023scannet++} datasets in mAP and mAP25, and mAP50 scores.}
  \label{tab:instance_segmentation}
  \centering
  \resizebox{\linewidth}{!}{
  \begin{tabular}{lccccccc}
    \toprule
    \multicolumn{1}{c}{\multirow{2}{*}{\textbf{Method}}} & \multicolumn{1}{c}{\multirow{2}{*}{\textbf{Venue}}} &\multicolumn{3}{c}{\textbf{ScanNet}} &\multicolumn{3}{c}{\textbf{ScanNet++}}\\
    \cmidrule(lr){3-5} \cmidrule(lr){6-8}
    \multicolumn{1}{c}{} & \multicolumn{1}{c}{} & \textbf{mAP} & \textbf{mAP$_{50}$} & \textbf{mAP$_{25}$} & \textbf{mAP} & \textbf{AP$_{50}$} & \textbf{AP$_{25}$}\\
    \midrule
    SAM3D~\cite{yang2023sam3d} & ICCV'23 & 13.7& 29.7& 54.5& 8.3& 17.5& 33.7\\
    SAM-graph~\cite{guo2023sam} & arXiv'23 & 15.1 & 33.3& 59.1& 12.9& 25.3& 43.6\\
    SAI3D~\cite{yin2023sai3d} & CVPR'24 & 18.8&42.5  & 62.3& 17.1& 31.1& 49.5\\
    SAMPro3D~\cite{xu2023sampro3d} & arXiv'23 & 22.2 &45.6  & 65.7& 18.9 & 33.7& 51.6\\
    PointSeg \cite{he2024pointseg} & arXiv'24 & 35.6 & 59.7 & 78.1 & 30.2 & 45.7 & 64.2\\
  \bottomrule
  \end{tabular}
  }
\end{table}

%% file: _tables/open_vocab.tex
\begin{table}[!ht]
\centering
\setlength{\tabcolsep}{3pt} 
\caption{Performance summary on ScanNet200 \cite{rozenberszki2022language}.
Three splits: Head, common, tail;
*: The metrics of mIoU (mAcc).}
\label{tab:open_vocab}
\resizebox{\linewidth}{!}{
    \begin{tabular}{llllllll}
        \toprule
        \textbf{Method} & \textbf{Venue} & \textbf{mAP} & \textbf{mAP$_{50}$} & \textbf{mAP$_{25}$} & \textbf{Head} & \textbf{Common} & \textbf{Tail} \\ 
        \midrule
        \multicolumn{8}{c}{\textit{Feature Distillation Methods}} \\
        \cmidrule(lr){2-5}
        OpenScene \cite{peng2023openscene} & CVPR'23 & 11.7 & 15.2 & 17.8 & 13.4 & 11.6 & 9.9 \\ 
        CLIP-FO3D \cite{zhang2023clip} & CVPR'23 & - & - & - & - & - & - \\ 
        \midrule
        \multicolumn{8}{c}{\textit{3D Captioning Methods}} \\
        \cmidrule(lr){2-5}
        PLA \cite{ding2022pla} & CVPR'23 & 1.8 (3.1)* & - & - & - & - & - \\ 
        RegionPLC \cite{yang2024regionplc} & CVPR'24 & 9.6 (17.8)* & - & - & - & - & - \\ 
        \midrule
        \multicolumn{8}{c}{\textit{3D projection Methods}} \\
        \cmidrule(lr){2-5}
        SemAbs \cite{hasemantic} & CoRL'22 & - & - & - & - & - & - \\ 
        OpenMask3D \cite{takmaz2023openmask3d} & NIPS'23 & 15.4 & 19.9 & 23.1 & 17.1 & 14.1 & 14.9 \\ 
        OpenIns3D \cite{huang2023openins3d} & ECCV'24 & 8.8 & 10.3 & 14.4 & 16 & 6.5 & 4.2 \\ 
        OVIR-3D \cite{lu2023ovir} & CoRL'23 & 13 & 24.9 & 32.3 & 14.4 & 12.7 & 11.7 \\ 
        Open3DIS \cite{nguyen2024open3dis} & CVPR'24 & 18.6 & 23.1 & 27.3 & 24.7 & 16.9 & 13.3 \\ 
        Segment3D \cite{huang2025segment3d} & ECCV'24 & - & - & - & - & - & - \\ 
        Open-YOLO3D \cite{boudjoghra2024open} & Arxiv'24 & 24.7 & 31.7 & 36.2 & 27.8 & 24.3 & 21.6 \\
    \bottomrule
    \end{tabular}
}
\end{table}

%% file: _sections/5_2d_vlms+llms+for_3d.tex
\section{Adapting 2DFMs + LLMs for 3D Tasks}
\label{sec:reasoning_tasks}
Recent advancements in LLMs have sparked interest in using these pre-trained models for 3D tasks. This approach helps overcome the scarcity of 3D datasets and annotations by leveraging dense descriptions and incorporating language as an additional modality to enhance 3D tasks. We now review both object-level and scene-level approaches, with the latter further categorised by architecture and the integration of 2DFMs and LLMs. In this section, we refer to large vision-language models (LVLMs) for 3D tasks as 3DLVLMs.

\subsection{Object Level 3DLVLMs}
\label{sec:object_llm}
This section will highlight the methods based on LLMs for object-level 3D tasks such as 3D object caption generation, retrieval, question answering, etc. 

We start by summarising a simplex method \textbf{Cap3D} \cite{luo2024scalable}, a method designed to generate descriptive captions for point clouds by leveraging the BLIP2 \cite{li2023blip}, CLIP \cite{radford2019language}, and GPT-4 \cite{achiam2023gpt} models. Cap3D employs a three-stage data processing pipeline to produce high-quality annotations: first, 3D assets are rendered into 2D images; these images are then passed through the BLIP2 model to generate initial captions. Next, the CLIP model filters these captions to ensure relevance and quality by measuring the alignment between the captions and the images. Finally, GPT-4 consolidates captions across multiple views of the same object, creating cohesive and contextually rich descriptions. This integrated approach harnesses the power of pre-trained models on large-scale text-image and text data to produce accurate and informative captions for 3D assets.


\textbf{X-InstructBLIP} \cite{panagopoulou2023x} integrates multiple modalities--such as images, text, audio, and video--into LLMs, enabling both single-modal reasoning tasks and cross-modal reasoning across three or more modalities. To achieve this, the paper explores two projection methods for frozen LLMs: Q-Formers \cite{li2023blip} and linear projection \cite{liu2024visual}. In response to the scarcity of instruction-tuning datasets for modalities beyond images, the authors developed a three-stage query-based data augmentation technique that leverages open-source LLMs to extract instruction-tuning data from existing captioning datasets. This approach allows for efficient dataset generation across diverse modalities. Extensive evaluations on 13 benchmarks spanning four modalities reveal that Q-Formers demonstrate strong performance in single-modal tasks and exhibit the versatility to switch flexibly between joint and discriminative reasoning in multi-modal contexts.

\textbf{GPT4Point} \cite{qi2024gpt4point} is a unified framework designed for point-language understanding and generation. It leverages a BERT-based Point-QFormer \cite{devlin2018bert, li2023blip} to align point-cloud features with textual descriptions, enabling deep integration of 3D geometry in text-based reasoning and generation tasks. This aligned feature representation is then fed into the model for text inference tasks and into a diffusion model for controlled 3D object generation for enhanced geometric and colour fidelity. To address the scarcity of annotated 3D point-language data, the work introduces Pyramid-XL, a data annotation engine built upon the Objaverse-XL \cite{deitke2024objaverse} dataset. Pyramid-XL uses VLMs to generate multi-view text annotations and organises these annotations into three hierarchical levels of detail, improving accuracy and completeness in 3D object descriptions. 

\textbf{ShapeLLM} \cite{qi2025shapellm} is a two-stage method aimed at supporting 3D tasks that require interaction with objects, such as object grasping. This method emphasizes multi-level shape understanding, addressing limitations in previous models that often rely on single-view 2D features, resulting in limited shape comprehension. Existing methods, such as ReCon \cite{qi2023contrast}, are hindered by scarce pre-training data. To overcome this, the authors introduce ReCon++, a 3D backbone that uses multi-view image tokens to capture semantic information across RGB and depth maps. They also propose a cross-modal alignment based on bipartite matching for implicit pose estimation and expand the pre-training dataset and parameters to improve 3D representations. A LLaMa-style finetuning \cite {touvron2023llama} is then applied to align 3D representations with language models.

\textbf{MiniGPT-3D} \cite{tang2024minigpt} 
addresses the challenges in resource-intensive VL alignment methods. Unlike previous approaches that require significant computational resources, MiniGPT-3D leverages 2D visual priors to efficiently bridge the gap between 2D and 3D modalities. It involves a four-stage training strategy that cascades through modality alignment steps, progressively refining the connection between point cloud data and language understanding. This method utilizes a mixture-of-expert model that adaptively aggregates features, enhancing efficiency without sacrificing accuracy. To further optimise performance, MiniGPT-3D employs 
LoRA \cite{hu2021lora} and Norm tuning. These techniques minimise the number of trainable parameters to just 47.8 million, a reduction of up to 260 times compared to other methods.

Different to MiniGPT-3D, \textbf{GreenPLM} \cite{tang2024more}, a model aimed at achieving 3D-language understanding with minimal 3D data by leveraging extensive text-based training to align with LLMs. Targeting the newly defined 3D data efficient point cloud understanding task, GreenPLM shifts the alignment focus from limited 3D data to abundant text data using the curated T3D dataset, which includes 6 million 3D object descriptions. Its methodology consists of a three-stage training process where the first two stages use only text data to build foundational language skills, while the third stage incorporates minimal 3D data to enhance point-LLMs alignment. Architecturally, GreenPLM employs 0M-Pooling, a cross-attention module that optimises token pooling, allowing the model to align point clouds with LLMs efficiently, even on limited hardware.

\textbf{LLaNA} \cite{amaduzzi2024llana} adopts a different approach by directly embedding neural radiation fields (NeRFs) in LLM for tasks such as NeRF \cite{mildenhall2021nerf} captioning, question-answering, and zero-shot classification. NeRFs, typically MLPs, capture both geometry and photorealistic details of objects, offering a richer alternative to traditional 2D images or point clouds. LLaNA utilises a meta-network encoder to process NeRF weights and map them into the embedding space of a pre-trained language model, specifically LLaMA 2 \cite{touvron2023llama}, allowing NeRF-language tasks to be performed without rendering NeRFs into 2D images. To train and evaluate LLaNA, the authors introduce a new NeRF-language dataset, generated with automated annotations based on LVLMs in NeRFs from ShapeNet \cite{chang_shapenet_2015}, supplemented by a split with manually curated descriptions, establishing a benchmark for NeRF-driven language tasks.

\subsection{Scene Level 3DLVLM}

\begin{figure*}[!htb]
  \centering
  \includegraphics[width=\linewidth]{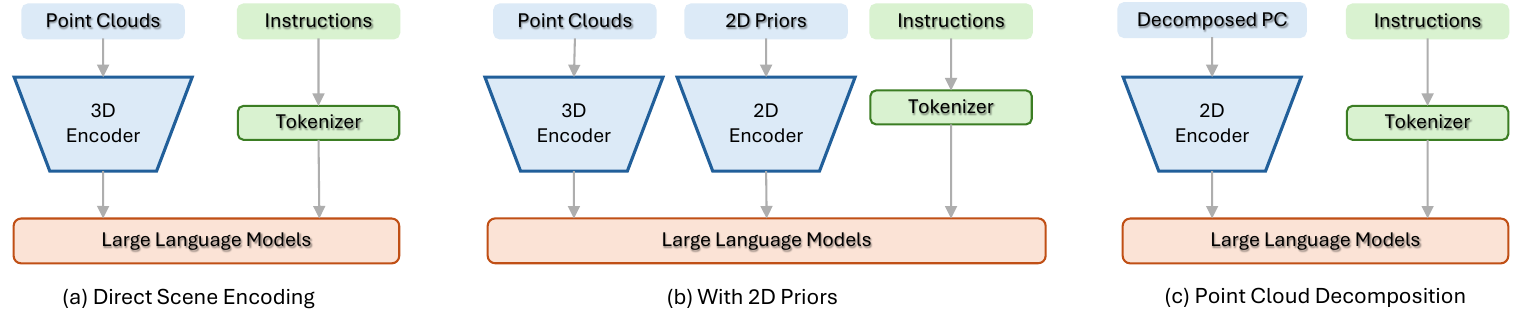}
  \caption{High level overview of scene level methods for 3D understanding, categorised by how point cloud data is handled.}
  \label{fig:3d_llm_scene}
\end{figure*}

\label{sec:scene_llm}
These methods focus on understanding complex real-world scenes using LVLMs. We categorise them into three groups based on how point cloud data is processed (Fig. \ref{fig:3d_llm_scene}): methods that directly handle point cloud scenes, those that leverage 2D priors like masks or features, and those that decompose point clouds into smaller assets or multi-view images.

\subsubsection{Direct Scene Encoding}
Large Language 3D Assistant (\textbf{LL3DA} \cite{chen2024ll3da}), showcases robust instruction-following capabilities in understanding, reasoning, and planning within complex 3D environments. By incorporating both textual instructions and potential visual interactions, LL3DA effectively reduces ambiguities in addressing various tasks across diverse 3D scenes. The model aggregates information from textual instructions, visual prompts, and 3D scenes into a fixed length of learnable querying tokens through an attention mechanism. These querying tokens act as a prefix for the textual instructions, enabling efficient interaction-aware 3D scene embeddings for instruction following.

\textbf{SIG3D} \cite{man2024situational} aims to improve the understanding and reasoning of embodied agents in complex 3D environments through effective situational modelling. It begins with estimating the agent’s ego-location and orientation using large-scale pre-trained language and visual encoders to enhance situational awareness from textual descriptions. Then tokenize the textual and 3D scene data, utilising a multi-modal transformer with self-attention and cross-attention mechanisms to fuse information from both modalities. To address the expansive search space in 3D environments, they reformulate the situation prediction task as an anchor-based classification problem, where visual tokens serve as anchor points, allowing for robust situational context estimation through regressed position and rotation parameters. Finally, situational alignment is applied to adjust and re-encode the visual tokens with situationally-aware embeddings, enhancing their relevance for 3D tasks.

\subsubsection{With 2D Priors}
A \textbf{RegionBLIP} \cite{zhou2023regionblip}, consisting of three core modules: modal feature extraction, alignment, and LLM comprehension. The extraction module retrieves features from modalities like images and point clouds, focussing on overall rather than fine-grained features, allowing shared encoders for both I-text and P-text data. For images, the frozen CLIP model is utilised, while the Point-BERT \cite{yu2022point} model is employed for point clouds. The alignment module is crucial, as LLMs are primarily trained on language data, making the comprehension of image or point cloud inputs challenging. By aligning features with textual descriptions through learnable queries, comprehension is enhanced before inputting them into the LLM, enabling effective processing of aligned features.

\textbf{3D-LLMs} \cite{hong20233d} emphasises the challenges of training from scratch due to limited datasets compared to the extensive billion-scale image-language datasets used for 2D FMs. Lacking pre-trained 3D encoders, they propose extracting 3D features from 2D multi-view images. This allows the use of pre-trained image encoders to map features to 3D data, integrating them into VLM backbones. The training process involves building aligned 3D features using three reconstruction methods: direct reconstruction, feature fusion, and neural field \cite{mildenhall2021nerf}. 3D-LLMs leverage frozen 2D VLMs, such as CLIP, for feature extraction. A 3D localisation mechanism is introduced to enhance spatial information absorption by augmenting 3D features with position embeddings and incorporating location tokens into the LLM vocabularies, effectively aligning 3D locations with the language model.

\textbf{LEO} \cite{huang2023embodied} is designed around two principles: it integrates multi-modal inputs--egocentric 2D images, global 3D data, and textual instructions--producing both textual responses and embodied action commands, and it leverages pre-trained LLMs for enhanced performance on downstream tasks. Data is converted into a token sequence for GPT-style autoregressive modeling \cite{brown2020language}. Tokenization uses SentencePiece for text, 2D image tokens for egocentric images, and object-centric 3D tokens from Mask3D proposals \cite{schult2023mask3d}. Continuous embodied actions are discretised for a unified action space. Various token embeddings are processed before being fed into the Vicuna-7B LLM \cite{vicuna2023}, which generates responses. Text and 2D tokens are embedded using lookup tables, while 3D tokens are refined with a Spatial Transformer for better 3D relationship capture. LoRA \cite{hu2021lora} is implemented to manage the alignment of multi-modal tokens while preserving the pre-trained model’s knowledge.

\textbf{SpatialRGPT} \cite{cheng2024spatialrgpt} discusses an approach to constructing a 3D scene graph from a single 2D image, focussing on open-vocabulary detection and segmentation, metric depth estimation, and camera calibration to generate accurate spatial representations. The model employs a visual encoder that integrates monocular depth information into an existing 2D VLM. The construction pipeline includes filtering unsuitable images, identifying objects, and creating a 3D scene graph composed of nodes and edges that represent object instances and their spatial relationships. To train the model, it generates spatially aware question-answer pairs using both template-based and LLM-based methods, resulting in a rich Open Spatial Dataset with millions of images and annotations. The SpatialRGPT architecture combines a visual encoder, a region-feature extractor, and a language model, incorporating depth information to enhance geometric reasoning while maintaining flexibility in handling RGB images and depth data.

\subsubsection{Point Cloud Decomposition}
The \textbf{Uni3D-LLM} \cite{liu2024uni3d} approach aligns point clouds with their corresponding images using modality-specific projectors. It employs a two-step process for point cloud alignment: first, objects are extracted with a detection method and then encoded using a pre-trained Point-BERT \cite{devlin2018bert} model, leveraging LLaMA2’s \cite{touvron2023llama} cognitive capabilities. For object-level tasks, point cloud data is mapped to textual space, while scene-level tasks utilize position encoding to preserve spatial integrity. Image alignment involves extracting features with multiple pre-trained encoders and addressing occlusion through a top-view representation. The LLM-to-Generation mapping block connects language model outputs to generation models via learnable generative tokens, guiding the diffusion process. The editing process allows modifications to the 3D model using rendered images, ensuring consistency through Gaussian splatting and instruct-pix2pix techniques. Training consists of two stages: first, focus on the text-to-generation mapping block and then on the perception module using PEFT \cite{xu2023parameter} with the Sphinx \cite{lin2023sphinx} model integrated with LLaMA2. 

Previous 2D LMMs utilise visual encoders to extract 2D features from images, aligning them with LLMs via projection layers for joint reasoning. \textbf{LLaVA-3D} \cite{zhu2024llava} enhances this by integrating 2D features into a 3D spatial context, creating 3D patches and employing 3D-aware pooling strategies to compress these patches, leading to a 3D-aware encoding and decoding process. Building on LLaVA \cite{liu2024visual}, which uses a CLIP encoder for 2D patch extraction, we introduce a 3D patch representation that embeds 3D spatial information into 2D features through a two-layer MLP. To manage computational overhead, we implement pooling mechanisms like voxelization pooling, which averages patches in occupied voxels, and farthest point sampling for representative subset selection. The final architecture enables the LLM to process 3D coordinates, generate outputs such as language responses and 3D bounding boxes, using a 3D Coordinate Token for coordinate context and a location token to guide accurate 3D box predictions, facilitating precise object localisation.

%% file: _sections/6_outlook_into_future.tex
\section{Outlook Into Future Directions}
\label{sec:outlook_into_future}

\textbf{Robust 3DFMs}
While there has been a significant surge in research aimed at developing 3DFMs by leveraging 2D FMs  (see Sec. \ref{sec:building_3dfms}), it is evident that there is significant room for improvement. These models demonstrate proficiency in understanding object-level point clouds; however, they often fall short when tasked with comprehending larger 3D point scenes. 
For example, models such as PointCLIP \cite{zhang2022pointclip} and PointMAE-I2P \cite{zhang2023learning} leverage 2DFMs to learn from point clouds. Yet, their downstream applications are primarily limited to small scenes from datasets such as PartNet, posing challenges when attempting to extend their applicability to larger scenes. 
Similarly, models such as ULIP/-V2 \cite{xue2023ulip} utilise 2D FMs to extract features from pre-trained 3D models, showcasing some advantages in tackling more complex 3D tasks. Despite these advancements, there remains a need to address the limitations associated with the scalability and generalisation of these models to larger and more diverse 3D environments.

\textbf{Large 3D datasets}
As we look toward the future of 3DFMs, it becomes increasingly apparent that addressing the scarcity of 3D data compared to the abundance of 2D image datasets is a pressing challenge. The question of how to bridge this gap is a pressing concern in the field. While the Internet is teeming with 2D images, the collection of 3D data remains a daunting task due to the high cost and time-consuming nature of equipment and processes involved in analysing and creating 3D datasets.
One potential solution lies in leveraging existing 2D datasets to generate 3D counterparts. While methods like ERFs \cite{mildenhall2021nerf} have been explored for this purpose, they often require significant computational resources and time to render a single scene. However, recent advancements in techniques such as Gaussian splatting \cite{kerbl3Dgaussians} offer a more computationally efficient approach. By harnessing these methods, researchers and practitioners can expedite the process of collecting large-scale 3D datasets, thus narrowing the gap between the availability of 2D and 3D data. For instance, datasets such as Objaverse \cite{objaverse} have demonstrated the potential of such approaches by providing a vast collection of 3D objects derived from various sources. These efforts exemplify how innovative methodologies can contribute to the rapid expansion of 3D data repositories, ultimately enhancing the development and applicability of future 3D foundational models.

\textbf{Efficient data and resources}
When FMs are applied to downstream tasks, they adapt efficiently with fewer parameters and data. However, they face challenges when dealing with complex datasets, such as 3D scenes in real-world environments \cite{ma2024llms}. To address these issues, we need more advanced methods that emulate human learning, such as adapting to minimal examples and learning from abstract concepts. Although these models show promising adaptability, deploying them on edge devices remains a significant hurdle \cite{zhu2023minigpt}. The few-shot learning approach aligns with this goal, but is currently limited to object-level scenarios and needs further development to handle more complex scene-level tasks \cite{ye2023closer, ahmadi2024foundation}.

\textbf{Continual adaptation}
FMs have been applied to various learning paradigms, including open-set and open-vocabulary tasks. However, their adaptability is often limited to single downstream datasets, lacking the flexibility to generalise across evolving data or new object classes in dynamic environments. In real-world 3D settings, where new data or object classes continuously emerge, adapting FMs to these changes remains a significant challenge \cite{yang2023geometry}. Although OV methods are useful, continual learning becomes essential in cases where OV is not feasible. For instance, where AI systems need to figure out class labels simply by observing objects or scenes, without relying on predefined descriptions \cite{lomonaco2020continual}. 

%% file: _sections/7_conclusion.tex
\section{Conclusion}
\label{sec:conclusion}
This paper presents a thorough examination of 2D image and large language pre-trained models, commonly referred to as Vision-Language Models (VLMs), Language Models (LLMs), and Foundation Models (FMs)--with a particular emphasis on their applications in 3D representation and understanding, especially regarding point cloud data. We highlight the transformative potential of FMs for enhancing 3D understanding. Key contributions include the categorisation of these methods based on various tasks and architectures. By establishing a well-structured taxonomy and providing concise introductions to various methodologies, we facilitate meaningful comparisons among them. Our novel grouping strategy and step-by-step explanations aim to cater to researchers at all levels, improving accessibility in this rapidly evolving field. This comprehensive overview not only outlines the current state of 3D representation techniques but also addresses the challenges faced in this domain and offers insights into future research directions, making it a valuable resource for guiding ongoing advancements in 3D understanding. 

%% file: main.bbl
\begin{thebibliography}{100}

\bibitem{heaven2016ai3d}
W.~D. Heaven.
\newblock AI begins to understand the 3-D world.
\newblock {\em MIT Technology Review}, December 9 2016.

\bibitem{lahoud20223d}
J.~Lahoud et~al.
\newblock 3D vision with transformers: A survey.
\newblock {\em arXiv preprint arXiv:2208.04309}, 2022.

\bibitem{wiki_pcloud}
W.~contributors.
\newblock Point cloud.
\newblock \url{https://en.wikipedia.org/wiki/Point_cloud}, 2025.
\newblock Accessed December 6, 2024.

\bibitem{qi2017pointnet}
C.~R. Qi et~al.
\newblock Pointnet: Deep learning on point sets for 3D classification and segmentation.
\newblock In {\em Proc. IEEE Conf. Comput. Vis. Pattern Recog. (CVPR)}, pp. 652--660, 2017.

\bibitem{lu2022transformers}
D.~Lu et~al.
\newblock Transformers in 3D point clouds: A survey.
\newblock {\em arXiv preprint arXiv:2205.07417}, 2022.

\bibitem{mao20233d}
J.~Mao et~al.
\newblock 3D object detection for autonomous driving: A comprehensive survey.
\newblock {\em Int. J. Comput. Vis. (IJCV)}, 131(8):1909--1963, 2023.

\bibitem{fei2023self}
B.~Fei et~al.
\newblock Self-supervised learning for pre-training 3D point clouds: A survey.
\newblock {\em arXiv preprint arXiv:2305.04691}, 2023.

\bibitem{dai2017scannet}
A.~Dai et~al.
\newblock ScanNet: Richly-annotated 3D Reconstructions of Indoor Scenes.
\newblock In {\em Proc. IEEE Conf. Comput. Vis. Pattern Recog. (CVPR)}, 2017.

\bibitem{yeshwanth2023scannet++}
C.~Yeshwanth et~al.
\newblock Scannet++: A high-fidelity dataset of 3D indoor scenes.
\newblock In {\em Proc. IEEE Int. Conf. Comput. Vis. (ICCV)}, pp. 12--22, 2023.

\bibitem{schult2023mask3d}
J.~Schult et~al.
\newblock Mask3d: Mask transformer for 3D semantic instance segmentation.
\newblock In {\em Proc. IEEE/RSJ Int. Conf. Robot. Autom. (ICRA)}, pp. 8216--8223. IEEE, 2023.

\bibitem{wang2023unibev}
S.~Wang et~al.
\newblock UniBEV: Multi-modal 3D Object Detection with Uniform BEV Encoders for Robustness against Missing Sensor Modalities.
\newblock In {\em IEEE Intelligent Vehicles Symposium}, 2024.

\bibitem{bommasani2021opportunities}
R.~Bommasani et~al.
\newblock On the opportunities and risks of foundation models.
\newblock {\em arXiv preprint arXiv:2108.07258}, 2021.

\bibitem{hinton2006fast}
G.~E. Hinton et~al.
\newblock A fast learning algorithm for deep belief nets.
\newblock {\em Neural computation}, 18(7):1527--1554, 2006.

\bibitem{perkins1999transfer}
D.~N. Perkins and G.~Salomon.
\newblock Transfer of Learning.
\newblock 1992.

\bibitem{devlin2018bert}
J.~Devlin et~al.
\newblock BERT: Pre-training of Deep Bidirectional Transformers for Language Understanding.
\newblock In {\em Proc. North Amer. Chapter Assoc. Comput. Linguistics (NAACL)}, pp. 4171--4186, 2019.

\bibitem{brown2020language}
T.~Brown et~al.
\newblock Language models are few-shot learners.
\newblock {\em Proc. Adv. Neural Inform. Process. Syst. (NeurIPS)}, 33:1877--1901, 2020.

\bibitem{radford2021learning}
A.~Radford et~al.
\newblock Learning transferable visual models from natural language supervision.
\newblock In {\em Proc. Int. Conf. Mach. Learn. (ICML)}, pp. 8748--8763, 2021.

\bibitem{zhang2022pointclipv2}
X.~Zhu et~al.
\newblock PointCLIP V2: Prompting CLIP and GPT for Powerful 3D Open-world Learning.
\newblock In {\em Proc. IEEE Int. Conf. Comput. Vis. (ICCV)}, pp. 2639--2650, 2023.

\bibitem{thengane2022clip}
V.~Thengane et~al.
\newblock Clip model is an efficient continual learner.
\newblock {\em arXiv preprint arXiv:2210.03114}, 2022.

\bibitem{kirillov2023segment}
A.~Kirillov et~al.
\newblock Segment anything.
\newblock {\em Proc. IEEE Int. Conf. Comput. Vis. (ICCV)}, 2023.

\bibitem{zhang2023segment}
Y.~Zhang and R.~Jiao.
\newblock How segment anything model (SAM) boost medical image segmentation?
\newblock {\em arXiv preprint arXiv:2305.03678}, 2023.

\bibitem{yang2023sam3d}
Y.~Yang et~al.
\newblock Sam3d: Segment anything in 3D scenes.
\newblock {\em Proc. IEEE Int. Conf. Comput. Vis. (ICCV)}, 2023.

\bibitem{xue2023ulip}
L.~Xue et~al.
\newblock ULIP: Learning a unified representation of language, images, and point clouds for 3D understanding.
\newblock In {\em Proc. IEEE Conf. Comput. Vis. Pattern Recog. (CVPR)}, pp. 1179--1189, 2023.

\bibitem{xue2023ulip2}
L.~Xue et~al.
\newblock ULIP-2: Towards Scalable Multimodal Pre-training For 3D Understanding.
\newblock {\em Proc. IEEE Conf. Comput. Vis. Pattern Recog. (CVPR)}, 2024.

\bibitem{zhang2022pointclip}
R.~Zhang et~al.
\newblock PointCLIP: Point Cloud Understanding by CLIP.
\newblock In {\em Proc. IEEE Conf. Comput. Vis. Pattern Recog. (CVPR)}, pp. 8552--8562, 2022.

\bibitem{liu2023partslip}
M.~Liu et~al.
\newblock Partslip: Low-shot part segmentation for 3D point clouds via pretrained image-language models.
\newblock In {\em Proc. IEEE Conf. Comput. Vis. Pattern Recog. (CVPR)}, pp. 21736--21746, 2023.

\bibitem{umam2023partdistill}
A.~Umam et~al.
\newblock PartDistill: 3D Shape Part Segmentation by Vision-Language Model Distillation.
\newblock {\em Proc. IEEE Conf. Comput. Vis. Pattern Recog. (CVPR)}, 2024.

\bibitem{bai2022transfusion}
X.~Bai et~al.
\newblock Transfusion: Robust LiDAR-camera fusion for 3D object detection with transformers.
\newblock In {\em Proc. IEEE Conf. Comput. Vis. Pattern Recog. (CVPR)}, pp. 1090--1099, 2022.

\bibitem{liu2023bevfusion}
Z.~Liu et~al.
\newblock Bevfusion: Multi-task multi-sensor fusion with unified bird's-eye view representation.
\newblock In {\em Proc. IEEE/RSJ Int. Conf. Robot. Autom. (ICRA)}, pp. 2774--2781. IEEE, 2023.

\bibitem{touvron2023llama}
H.~Touvron et~al.
\newblock Llama 2: Open foundation and fine-tuned chat models.
\newblock {\em arXiv preprint arXiv:2307.09288}, 2023.

\bibitem{vicuna2023}
W.-L. Chiang et~al.
\newblock Vicuna: An Open-Source Chatbot Impressing GPT-4 with 90\%* ChatGPT Quality, 2023.

\bibitem{gunasekar2023textbooks}
S.~Gunasekar et~al.
\newblock Textbooks are all you need.
\newblock {\em arXiv preprint arXiv:2306.11644}, 2023.

\bibitem{luo2024scalable}
T.~Luo et~al.
\newblock Scalable 3D captioning with pretrained models.
\newblock {\em Proc. Adv. Neural Inform. Process. Syst. (NeurIPS)}, 36, 2024.

\bibitem{qi2025shapellm}
Z.~Qi et~al.
\newblock Shapellm: Universal 3D object understanding for embodied interaction.
\newblock In {\em Proc. Eur. Conf. Comput. Vis. (ECCV)}, pp. 214--238. Springer, 2025.

\bibitem{zhou2023regionblip}
Q.~Zhou et~al.
\newblock Regionblip: A unified multi-modal pre-training framework for holistic and regional comprehension.
\newblock {\em arXiv preprint arXiv:2308.02299}, 2023.

\bibitem{chen2024ll3da}
S.~Chen et~al.
\newblock LL3DA: Visual Interactive Instruction Tuning for Omni-3D Understanding Reasoning and Planning.
\newblock In {\em Proc. IEEE Conf. Comput. Vis. Pattern Recog. (CVPR)}, pp. 26428--26438, 2024.

\bibitem{roh2022languagerefer}
J.~Roh et~al.
\newblock Languagerefer: Spatial-language model for 3D visual grounding.
\newblock In {\em Conference on Robot Learning}, pp. 1046--1056, 2022.

\bibitem{huang2023embodied}
J.~Huang et~al.
\newblock An embodied generalist agent in 3D world.
\newblock {\em Proc. Int. Conf. Mach. Learn. (ICML)}, 2024.

\bibitem{9965773}
S.~Ye et~al.
\newblock 3D Question Answering.
\newblock {\em IEEE Trans. Vis. Comput. Graph. (TVCG)}, pp. 1--16, 2022.

\bibitem{han2023imagebind}
J.~Han et~al.
\newblock Imagebind-LLM: Multi-modality instruction tuning.
\newblock {\em arXiv preprint arXiv:2309.03905}, 2023.

\bibitem{jia2021scaling}
C.~Jia et~al.
\newblock Scaling up visual and vision-language representation learning with noisy text supervision.
\newblock In {\em Proc. Int. Conf. Mach. Learn. (ICML)}, pp. 4904--4916, 2021.

\bibitem{liu2024visual}
H.~Liu et~al.
\newblock Visual instruction tuning.
\newblock {\em Proc. Adv. Neural Inform. Process. Syst. (NeurIPS)}, 36, 2024.

\bibitem{zhu2023minigpt}
D.~Zhu et~al.
\newblock Minigpt-4: Enhancing vision-language understanding with advanced large language models.
\newblock {\em Proc. Int. Conf. Learn. Represent. (ICLR)}, 2024.

\bibitem{guo2020deep}
Y.~Guo et~al.
\newblock Deep learning for 3D point clouds: A survey.
\newblock {\em IEEE Trans. Pattern Anal. Mach. Intell. (TPAMI)}, 43(12):4338--4364, 2020.

\bibitem{zeng2022survey}
J.~Zeng et~al.
\newblock A survey on transformers for point cloud processing: An updated overview.
\newblock {\em IEEE Access}, 10:86510--86527, 2022.

\bibitem{wang2023multi}
Y.~Wang et~al.
\newblock Multi-modal 3D object detection in autonomous driving: A survey.
\newblock {\em Int. J. Comput. Vis. (IJCV)}, 131(8):2122--2152, 2023.

\bibitem{prakash2021multi}
A.~Prakash et~al.
\newblock Multi-modal fusion transformer for end-to-end autonomous driving.
\newblock In {\em Proc. IEEE Conf. Comput. Vis. Pattern Recog. (CVPR)}, pp. 7077--7087, 2021.

\bibitem{tang2023multi}
Y.~Tang et~al.
\newblock Multi-modality 3D object detection in autonomous driving: A review.
\newblock {\em Neurocomputing}, pp. 126587, 2023.

\bibitem{awais2023foundational}
M.~Awais et~al.
\newblock Foundational models defining a new era in vision: A survey and outlook.
\newblock {\em IEEE Trans. Pattern Anal. Mach. Intell. (TPAMI)}, 2024.

\bibitem{xiao2023survey}
A.~Xiao et~al.
\newblock A survey of label-efficient deep learning for 3D point clouds.
\newblock {\em IEEE Trans. Pattern Anal. Mach. Intell. (TPAMI)}, 2024.

\bibitem{chang_shapenet_2015}
A.~X. Chang et~al.
\newblock {ShapeNet}: An Information-Rich 3D Model Repository.

\bibitem{Wu_2015_CVPR}
Z.~Wu et~al.
\newblock 3D ShapeNets: A Deep Representation for Volumetric Shapes.
\newblock In {\em Proc. IEEE Conf. Comput. Vis. Pattern Recog. (CVPR)}, 2015.

\bibitem{fu20213d}
H.~Fu et~al.
\newblock 3D-future: 3D furniture shape with texture.
\newblock {\em Int. J. Comput. Vis. (IJCV)}, 129:3313--3337, 2021.

\bibitem{collins2022abo}
J.~Collins et~al.
\newblock Abo: Dataset and benchmarks for real-world 3D object understanding.
\newblock In {\em Proc. IEEE Conf. Comput. Vis. Pattern Recog. (CVPR)}, pp. 21126--21136, 2022.

\bibitem{Jensen_2014_CVPR}
R.~Jensen et~al.
\newblock Large Scale Multi-view Stereopsis Evaluation.
\newblock In {\em Proc. IEEE Conf. Comput. Vis. Pattern Recog. (CVPR)}, 2014.

\bibitem{yao2020blendedmvs}
Y.~Yao et~al.
\newblock Blendedmvs: A large-scale dataset for generalized multi-view stereo networks.
\newblock In {\em Proc. IEEE Conf. Comput. Vis. Pattern Recog. (CVPR)}, pp. 1790--1799, 2020.

\bibitem{uy2029scanobjectnn}
M.~A. Uy et~al.
\newblock Revisiting Point Cloud Classification: A New Benchmark Dataset and Classification Model on Real-World Data.
\newblock In {\em International Conference on Computer Vision (ICCV)}, 2019.

\bibitem{downs2022google}
L.~Downs et~al.
\newblock Google scanned objects: A high-quality dataset of 3D scanned household items.
\newblock In {\em Proc. IEEE/RSJ Int. Conf. Robot. Autom. (ICRA)}, pp. 2553--2560. IEEE, 2022.

\bibitem{liu2022akb}
L.~Liu et~al.
\newblock Akb-48: A real-world articulated object knowledge base.
\newblock In {\em Proc. IEEE Conf. Comput. Vis. Pattern Recog. (CVPR)}, pp. 14809--14818, 2022.

\bibitem{reizenstein2021common}
J.~Reizenstein et~al.
\newblock Common objects in 3D: Large-scale learning and evaluation of real-life 3D category reconstruction.
\newblock In {\em Proc. IEEE Int. Conf. Comput. Vis. (ICCV)}, pp. 10901--10911, 2021.

\bibitem{wu2023omniobject3d}
T.~Wu et~al.
\newblock Omniobject3d: Large-vocabulary 3D object dataset for realistic perception, reconstruction and generation.
\newblock In {\em Proc. IEEE Conf. Comput. Vis. Pattern Recog. (CVPR)}, pp. 803--814, 2023.

\bibitem{Matterport3D}
A.~Chang et~al.
\newblock {Matterport3D}: Learning from {RGB-D} Data in Indoor Environments.
\newblock {\em Proc. IEEE Int. Conf. 3D Vision (3DV)}, 2017.

\bibitem{zhou2018stereo}
T.~Zhou et~al.
\newblock Stereo magnification: Learning view synthesis using multiplane images.
\newblock {\em Proc. ACM Conf. Comput. Graph. (SIGGRAPH)}, 2018.

\bibitem{baruch2021arkitscenes}
G.~Baruch et~al.
\newblock Arkitscenes: A diverse real-world dataset for 3D indoor scene understanding using mobile rgb-d data.
\newblock {\em Proc. Adv. Neural Inform. Process. Syst. (NeurIPS)}, 2021.

\bibitem{rozenberszki2022language}
D.~Rozenberszki et~al.
\newblock Language-Grounded Indoor 3D Semantic Segmentation in the Wild.
\newblock In {\em Proceedings of the European Conference on Computer Vision ({ECCV})}, 2022.

\bibitem{behley2019iccv}
J.~Behley et~al.
\newblock SemanticKITTI: A Dataset for Semantic Scene Understanding of LiDAR Sequences.
\newblock In {\em Proc. IEEE Int. Conf. Comput. Vis. (ICCV)}, 2019.

\bibitem{deschaud2021paris}
J.-E. Deschaud et~al.
\newblock Paris-CARLA-3D: A real and synthetic outdoor point cloud dataset for challenging tasks in 3D mapping.
\newblock {\em Remote Sensing}, 13(22):4713, 2021.

\bibitem{caesar2020nuscenes}
H.~Caesar et~al.
\newblock nuscenes: A multimodal dataset for autonomous driving.
\newblock In {\em Proc. IEEE Conf. Comput. Vis. Pattern Recog. (CVPR)}, pp. 11621--11631, 2020.

\bibitem{dai_scannet_2017}
A.~Dai et~al.
\newblock {ScanNet}: Richly-Annotated 3D Reconstructions of Indoor Scenes.
\newblock In {\em Proc. IEEE Conf. Comput. Vis. Pattern Recog. (CVPR)}, pp. 5828--5839, 2017.

\bibitem{caesar_nuscenes_2020}
H.~Caesar et~al.
\newblock {nuScenes}: A Multimodal Dataset for Autonomous Driving.
\newblock In {\em Proc. IEEE Conf. Comput. Vis. Pattern Recog. (CVPR)}, pp. 11621--11631, 2020.

\bibitem{deitke2023objaverse}
M.~Deitke et~al.
\newblock Objaverse: A universe of annotated 3D objects.
\newblock In {\em Proc. IEEE Conf. Comput. Vis. Pattern Recog. (CVPR)}, pp. 13142--13153, 2023.

\bibitem{chen2018text2shape}
K.~Chen et~al.
\newblock Text2Shape: Generating Shapes from Natural Language by Learning Joint Embeddings.
\newblock {\em Proc. Asian Conf. Comput. Vis. (ACCV)}, 2019.

\bibitem{jia2024sceneverse}
B.~Jia et~al.
\newblock SceneVerse: Scaling 3D Vision-Language Learning for Grounded Scene Understanding.
\newblock {\em arXiv preprint arXiv:2401.09340}, 2024.

\bibitem{yang2023lidar}
S.~Yang et~al.
\newblock LiDAR-LLM: Exploring the potential of large language models for 3D LiDAR understanding.
\newblock {\em arXiv preprint arXiv:2312.14074}, 2023.

\bibitem{achiam2023gpt}
J.~Achiam et~al.
\newblock Gpt-4 technical report.
\newblock {\em arXiv preprint arXiv:2303.08774}, 2023.

\bibitem{chen2020scanrefer}
D.~Z. Chen et~al.
\newblock ScanRefer: 3D Object Localization in RGB-D Scans using Natural Language.
\newblock {\em Proc. Eur. Conf. Comput. Vis. (ECCV)}, 2020.

\bibitem{achlioptas2020referit_3d}
P.~Achlioptas et~al.
\newblock {ReferIt3D}: Neural Listeners for Fine-Grained 3D Object Identification in Real-World Scenes.
\newblock In {\em Proc. Eur. Conf. Comput. Vis. (ECCV)}, 2020.

\bibitem{zhang2023multi3drefer}
Y.~Zhang et~al.
\newblock Multi3drefer: Grounding text description to multiple 3D objects.
\newblock In {\em Proc. IEEE Int. Conf. Comput. Vis. (ICCV)}, pp. 15225--15236, 2023.

\bibitem{ma2024llms}
X.~Ma et~al.
\newblock When LLMs step into the 3D World: A survey and Meta-Analysis of 3D Tasks via Multi-modal Large Language Models.
\newblock {\em arXiv preprint arXiv:2405.10255}, 2024.

\bibitem{donahue2014decaf}
J.~Donahue et~al.
\newblock Decaf: A deep convolutional activation feature for generic visual recognition.
\newblock In {\em Proc. Int. Conf. Mach. Learn. (ICML)}, pp. 647--655, 2014.

\bibitem{russakovsky2015imagenet}
O.~Russakovsky et~al.
\newblock Imagenet large scale visual recognition challenge.
\newblock {\em Int. J. Comput. Vis. (IJCV)}, 115:211--252, 2015.

\bibitem{he2016deep}
K.~He et~al.
\newblock Deep residual learning for image recognition.
\newblock In {\em Proc. IEEE Conf. Comput. Vis. Pattern Recog. (CVPR)}, pp. 770--778, 2015.

\bibitem{radford2019language}
A.~Radford et~al.
\newblock Language models are unsupervised multitask learners.
\newblock {\em OpenAI blog}, 1(8):9, 2019.

\bibitem{wei2022emergent}
J.~Wei et~al.
\newblock Emergent abilities of large language models.
\newblock {\em arXiv preprint arXiv:2206.07682}, 2022.

\bibitem{dosovitskiy2020image}
A.~Dosovitskiy et~al.
\newblock An image is worth 16x16 words: Transformers for image recognition at scale.
\newblock {\em Proc. Int. Conf. Learn. Represent. (ICLR)}, 2021.

\bibitem{he2022masked}
K.~He et~al.
\newblock Masked autoencoders are scalable vision learners.
\newblock In {\em Proc. IEEE Conf. Comput. Vis. Pattern Recog. (CVPR)}, pp. 16000--16009, 2022.

\bibitem{deng2009imagenet}
J.~Deng et~al.
\newblock Imagenet: A large-scale hierarchical image database.
\newblock In {\em Proc. IEEE Conf. Comput. Vis. Pattern Recog. (CVPR)}, pp. 248--255. Ieee, 2009.

\bibitem{sun2017revisiting}
C.~Sun et~al.
\newblock Revisiting unreasonable effectiveness of data in deep learning era.
\newblock In {\em Proc. IEEE Int. Conf. Comput. Vis. (ICCV)}, 2017.

\bibitem{zhao2023survey}
W.~X. Zhao et~al.
\newblock A survey of large language models.
\newblock {\em arXiv preprint arXiv:2303.18223}, 2023.

\bibitem{li2023blip}
J.~Li et~al.
\newblock Blip-2: Bootstrapping language-image pre-training with frozen image encoders and large language models.
\newblock In {\em Proc. Int. Conf. Mach. Learn. (ICML)}, pp. 19730--19742, 2023.

\bibitem{alayrac2022flamingo}
J.-B. Alayrac et~al.
\newblock Flamingo: a visual language model for few-shot learning.
\newblock {\em Proc. Adv. Neural Inform. Process. Syst. (NeurIPS)}, 35:23716--23736, 2022.

\bibitem{xu2023parameter}
L.~Xu et~al.
\newblock Parameter-efficient fine-tuning methods for pretrained language models: A critical review and assessment.
\newblock {\em arXiv preprint arXiv:2312.12148}, 2023.

\bibitem{wu2023medical}
J.~Wu et~al.
\newblock Medical sam adapter: Adapting segment anything model for medical image segmentation.
\newblock {\em arXiv preprint arXiv:2304.12620}, 2023.

\bibitem{lester2021power}
B.~Lester et~al.
\newblock The power of scale for parameter-efficient prompt tuning.
\newblock {\em Conf. Empir. Methods Nat. Lang. Process. (EMNLP)}, 2021.

\bibitem{hu2021lora}
E.~J. Hu et~al.
\newblock Lora: Low-rank adaptation of large language models.
\newblock {\em Proc. Int. Conf. Learn. Represent. (ICLR)}, 2022.

\bibitem{dettmers2024qlora}
T.~Dettmers et~al.
\newblock Qlora: Efficient finetuning of quantized llms.
\newblock {\em Proc. Adv. Neural Inform. Process. Syst. (NeurIPS)}, 36, 2024.

\bibitem{xu2021image2point}
C.~Xu et~al.
\newblock Image2point: 3D point-cloud understanding with 2d image pretrained models.
\newblock {\em Proc. Eur. Conf. Comput. Vis. (ECCV)}, 2022.

\bibitem{qian2022improving}
G.~Qian et~al.
\newblock Pix4Point: Image Pretrained Standard Transformers for 3D Point Cloud Understanding.
\newblock {\em Proc. IEEE Int. Conf. 3D Vision (3DV)}, 2024.

\bibitem{kang2023point}
J.~Kang et~al.
\newblock Point Clouds Are Specialized Images: A Knowledge Transfer Approach for 3D Understanding.
\newblock {\em IEEE Trans. Multimedia (TMM)}, 2024.

\bibitem{wu_3d_2015}
Z.~Wu et~al.
\newblock {3D {ShapeNets}: A Deep Representation for Volumetric Shapes}.
\newblock In {\em Proc. IEEE Conf. Comput. Vis. Pattern Recog. (CVPR)}, pp. 1912--1920, 2015.

\bibitem{shen2023diffclip}
S.~Shen et~al.
\newblock DiffCLIP: Leveraging Stable Diffusion for Language Grounded 3D Classification.
\newblock {\em Proc. IEEE Winter Conf. Appl. Comput. Vis. (WACV)}, 2024.

\bibitem{zhang2023adding}
L.~Zhang et~al.
\newblock Adding conditional control to text-to-image diffusion models.
\newblock In {\em Proc. IEEE Int. Conf. Comput. Vis. (ICCV)}, pp. 3836--3847, 2023.

\bibitem{rombach2022high}
R.~Rombach et~al.
\newblock High-resolution image synthesis with latent diffusion models.
\newblock In {\em Proc. IEEE Conf. Comput. Vis. Pattern Recog. (CVPR)}, pp. 10684--10695, 2022.

\bibitem{schuhmann2022laion}
C.~Schuhmann et~al.
\newblock Laion-5b: An open large-scale dataset for training next generation image-text models.
\newblock {\em Proc. Adv. Neural Inform. Process. Syst. (NeurIPS)}, 35:25278--25294, 2022.

\bibitem{liu2021learning}
Y.-C. Liu et~al.
\newblock Learning from 2d: Contrastive pixel-to-point knowledge transfer for 3D pretraining.
\newblock {\em arXiv preprint arXiv:2104.04687}, 2021.

\bibitem{choy20194d}
C.~Choy et~al.
\newblock 4d spatio-temporal convnets: Minkowski convolutional neural networks.
\newblock In {\em Proc. IEEE Conf. Comput. Vis. Pattern Recog. (CVPR)}, pp. 3075--3084, 2019.

\bibitem{afham2022crosspoint}
M.~Afham et~al.
\newblock Crosspoint: Self-supervised cross-modal contrastive learning for 3D point cloud understanding.
\newblock In {\em Proc. IEEE Conf. Comput. Vis. Pattern Recog. (CVPR)}, pp. 9902--9912, 2022.

\bibitem{wang2019dynamic}
Y.~Wang et~al.
\newblock Dynamic graph cnn for learning on point clouds.
\newblock {\em ACM Transactions on Graphics (tog)}, 38(5):1--12, 2019.

\bibitem{huang2023clip2point}
T.~Huang et~al.
\newblock Clip2point: Transfer clip to point cloud classification with image-depth pre-training.
\newblock In {\em Proc. IEEE Int. Conf. Comput. Vis. (ICCV)}, pp. 22157--22167, 2023.

\bibitem{vaswani2017attention}
A.~Vaswani et~al.
\newblock Attention is all you need.
\newblock {\em Proc. Adv. Neural Inform. Process. Syst. (NeurIPS)}, 30, 2017.

\bibitem{yao20223d}
Y.~Yao et~al.
\newblock 3D Point Cloud Pre-training with Knowledge Distillation from 2D Images.
\newblock {\em arXiv preprint arXiv:2212.08974}, 2022.

\bibitem{mokady2021clipcap}
R.~Mokady et~al.
\newblock Clipcap: Clip prefix for image captioning.
\newblock {\em arXiv preprint arXiv:2111.09734}, 2021.

\bibitem{zhang2023learning}
R.~Zhang et~al.
\newblock Learning 3D representations from 2d pre-trained models via image-to-point masked autoencoders.
\newblock In {\em Proc. IEEE Conf. Comput. Vis. Pattern Recog. (CVPR)}, pp. 21769--21780, 2023.

\bibitem{chen2023bridging}
Z.~Chen and B.~Li.
\newblock Bridging the Domain Gap: Self-Supervised 3D Scene Understanding with Foundation Models.
\newblock {\em Proc. Adv. Neural Inform. Process. Syst. (NeurIPS)}, 2023.

\bibitem{oquab2023dinov2}
M.~Oquab et~al.
\newblock Dinov2: Learning robust visual features without supervision.
\newblock {\em Trans. Mach. Learn. Res (TMLR)}, 2024.

\bibitem{huang2023tag2text}
X.~Huang et~al.
\newblock Tag2text: Guiding vision-language model via image tagging.
\newblock {\em Proc. Int. Conf. Learn. Represent. (ICLR)}, 2024.

\bibitem{song2015sunrgbd}
S.~Song et~al.
\newblock SUN RGB-D: A RGB-D scene understanding benchmark suite.
\newblock In {\em Proc. IEEE Conf. Comput. Vis. Pattern Recog. (CVPR)}, pp. 567--576, 2015.

\bibitem{yu2022point}
X.~Yu et~al.
\newblock Point-bert: Pre-training 3D point cloud transformers with masked point modeling.
\newblock In {\em Proc. IEEE Conf. Comput. Vis. Pattern Recog. (CVPR)}, pp. 19313--19322, 2022.

\bibitem{qian2022pointnext}
G.~Qian et~al.
\newblock Pointnext: Revisiting pointnet++ with improved training and scaling strategies.
\newblock {\em Proc. Adv. Neural Inform. Process. Syst. (NeurIPS)}, 35:23192--23204, 2022.

\bibitem{hegde2023clip}
D.~Hegde et~al.
\newblock Clip goes 3D: Leveraging prompt tuning for language grounded 3D recognition.
\newblock In {\em Proc. IEEE Int. Conf. Comput. Vis. (ICCV)}, pp. 2028--2038, 2023.

\bibitem{zhao2021point}
H.~Zhao et~al.
\newblock Point transformer.
\newblock In {\em Proc. IEEE Int. Conf. Comput. Vis. (ICCV)}, pp. 16259--16268, 2021.

\bibitem{ma2022rethinking}
X.~Ma et~al.
\newblock Rethinking network design and local geometry in point cloud: A simple residual MLP framework.
\newblock {\em Proc. Int. Conf. Learn. Represent. (ICLR)}, 2022.

\bibitem{chen2023clip2scene}
R.~Chen et~al.
\newblock CLIP2Scene: Towards Label-efficient 3D Scene Understanding by CLIP.
\newblock In {\em Proc. IEEE Conf. Comput. Vis. Pattern Recog. (CVPR)}, pp. 7020--7030, 2023.

\bibitem{tang2020searching}
H.~Tang et~al.
\newblock Searching efficient 3D architectures with sparse point-voxel convolution.
\newblock In {\em Proc. Eur. Conf. Comput. Vis. (ECCV)}, pp. 685--702. Springer, 2020.

\bibitem{huang2023joint}
R.~Huang et~al.
\newblock Joint representation learning for text and 3D point cloud.
\newblock {\em Pattern Recognition}, pp. 110086, 2023.

\bibitem{zeng2023clip2}
Y.~Zeng et~al.
\newblock CLIP2: Contrastive Language-Image-Point Pretraining from Real-World Point Cloud Data.
\newblock In {\em Proc. IEEE Conf. Comput. Vis. Pattern Recog. (CVPR)}, pp. 15244--15253, 2023.

\bibitem{qi2017pointnet++}
C.~R. Qi et~al.
\newblock Pointnet++: Deep hierarchical feature learning on point sets in a metric space.
\newblock {\em Proc. Adv. Neural Inform. Process. Syst. (NeurIPS)}, 30, 2017.

\bibitem{liu2023openshape}
M.~Liu et~al.
\newblock OpenShape: Scaling Up 3D Shape Representation Towards Open-World Understanding.
\newblock {\em Proc. Adv. Neural Inform. Process. Syst. (NeurIPS)}, 2024.

\bibitem{delitzas2023multi}
A.~Delitzas et~al.
\newblock Multi-CLIP: Contrastive Vision-Language Pre-training for Question Answering tasks in 3D Scenes.
\newblock {\em Proc. Brit. Mach. Vis. Conf. (BMVC)}, 2023.

\bibitem{zhou2023uni3d}
J.~Zhou et~al.
\newblock Uni3d: Exploring unified 3D representation at scale.
\newblock {\em Proc. Int. Conf. Learn. Represent. (ICLR)}, 2024.

\bibitem{fang2023eva}
Y.~Fang et~al.
\newblock Eva: Exploring the limits of masked visual representation learning at scale.
\newblock In {\em Proc. IEEE Conf. Comput. Vis. Pattern Recog. (CVPR)}, pp. 19358--19369, 2023.

\bibitem{caron2021emerging}
M.~Caron et~al.
\newblock Emerging properties in self-supervised vision transformers.
\newblock In {\em Proc. IEEE Int. Conf. Comput. Vis. (ICCV)}, pp. 9650--9660, 2021.

\bibitem{ji2023jm3d}
J.~Ji et~al.
\newblock JM3D \& JM3D-LLM: Elevating 3D Representation with Joint Multi-modal Cues.
\newblock {\em arXiv preprint arXiv:2310.09503}, 2023.

\bibitem{xie2020pointcontrast}
S.~Xie et~al.
\newblock Pointcontrast: Unsupervised pre-training for 3D point cloud understanding.
\newblock In {\em Proc. Eur. Conf. Comput. Vis. (ECCV)}, pp. 574--591. Springer, 2020.

\bibitem{wang2021unsupervised}
H.~Wang et~al.
\newblock Unsupervised point cloud pre-training via occlusion completion.
\newblock In {\em Proc. IEEE Int. Conf. Comput. Vis. (ICCV)}, pp. 9782--9792, 2021.

\bibitem{hou2021exploring}
J.~Hou et~al.
\newblock Exploring data-efficient 3D scene understanding with contrastive scene contexts.
\newblock In {\em Proc. IEEE Conf. Comput. Vis. Pattern Recog. (CVPR)}, pp. 15587--15597, 2021.

\bibitem{Zhang_2021_ICCV}
Z.~Zhang et~al.
\newblock Self-Supervised Pretraining of 3D Features on Any Point-Cloud.
\newblock In {\em Proc. IEEE Int. Conf. Comput. Vis. (ICCV)}, pp. 10252--10263, 2021.

\bibitem{Yu_2022_CVPR}
X.~Yu et~al.
\newblock Point-BERT: Pre-Training 3D Point Cloud Transformers With Masked Point Modeling.
\newblock In {\em Proc. IEEE Conf. Comput. Vis. Pattern Recog. (CVPR)}, pp. 19313--19322, 2022.

\bibitem{zhang2022point}
R.~Zhang et~al.
\newblock Point-m2ae: multi-scale masked autoencoders for hierarchical point cloud pre-training.
\newblock {\em Proc. Adv. Neural Inform. Process. Syst. (NeurIPS)}, 35:27061--27074, 2022.

\bibitem{hess2022masked}
G.~Hess et~al.
\newblock Masked autoencoders for self-supervised learning on automotive point clouds.
\newblock {\em Proc. Eur. Conf. Comput. Vis. (ECCV)}, 2022.

\bibitem{li2022simipu}
Z.~Li et~al.
\newblock Simipu: Simple 2d image and 3D point cloud unsupervised pre-training for spatial-aware visual representations.
\newblock In {\em Proc. AAAI Conf. Artif. Intell. (AAAI)}, volume~36, pp. 1500--1508, 2022.

\bibitem{bao2021vlmo}
H.~Bao et~al.
\newblock Vlmo: Unified vision-language pre-training with mixture-of-modality-experts.
\newblock {\em arXiv preprint arXiv:2111.02358}, 2021.

\bibitem{Hadsell2006dimensionality}
R.~Hadsell et~al.
\newblock Dimensionality Reduction by Learning an Invariant Mapping.
\newblock In {\em Proc. IEEE Conf. Comput. Vis. Pattern Recog. (CVPR)}, volume~2, pp. 1735--1742, 2006.

\bibitem{gutmann2010noise}
M.~Gutmann and A.~Hyvärinen.
\newblock Noise-contrastive estimation: A new estimation principle for unnormalized statistical models.
\newblock In Y.~W. Teh and M.~Titterington, editors, {\em Proc. Int. Conf. Artif. Intell. Statist. (AISTAT)}, volume~9 of {\em Proceedings of Machine Learning Research}, pp. 297--304, Chia Laguna Resort, Sardinia, Italy, 13--15 May 2010. PMLR.

\bibitem{chen2020simple}
T.~Chen et~al.
\newblock A simple framework for contrastive learning of visual representations.
\newblock In {\em Proc. Int. Conf. Mach. Learn. (ICML)}, pp. 1597--1607, 2020.

\bibitem{oord2018representation}
A.~v.~d. Oord et~al.
\newblock Representation learning with contrastive predictive coding.
\newblock {\em arXiv preprint arXiv:1807.03748}, 2018.

\bibitem{hinton2015distilling}
G.~Hinton.
\newblock Distilling the Knowledge in a Neural Network.
\newblock {\em arXiv preprint arXiv:1503.02531}, 2015.

\bibitem{huang2024joint}
R.~Huang et~al.
\newblock Joint representation learning for text and 3D point cloud.
\newblock {\em Pattern Recognition}, 147:110086, 2024.

\bibitem{cherti2023reproducible}
M.~Cherti et~al.
\newblock Reproducible scaling laws for contrastive language-image learning.
\newblock In {\em Proc. IEEE Conf. Comput. Vis. Pattern Recog. (CVPR)}, pp. 2818--2829, 2023.

\bibitem{wang2022p2p}
Z.~Wang et~al.
\newblock P2p: Tuning pre-trained image models for point cloud analysis with point-to-pixel prompting.
\newblock {\em Proc. Adv. Neural Inform. Process. Syst. (NeurIPS)}, 35:14388--14402, 2022.

\bibitem{guo2023calip}
Z.~Guo et~al.
\newblock Calip: Zero-shot enhancement of clip with parameter-free attention.
\newblock In {\em Proc. AAAI Conf. Artif. Intell. (AAAI)}, volume~37, pp. 746--754, 2023.

\bibitem{peng2023multi}
H.~Peng et~al.
\newblock Multi-view Vision-Prompt Fusion Network: Can 2D Pre-trained Model Boost 3D Point Cloud Data-scarce Learning?
\newblock {\em arXiv preprint arXiv:2304.10224}, 2023.

\bibitem{yi2023invariant}
X.~Yi et~al.
\newblock Invariant training 2d-3D joint hard samples for few-shot point cloud recognition.
\newblock In {\em Proc. IEEE Int. Conf. Comput. Vis. (ICCV)}, pp. 14463--14474, 2023.

\bibitem{jia2022visual}
M.~Jia et~al.
\newblock Visual prompt tuning.
\newblock In {\em Proc. Eur. Conf. Comput. Vis. (ECCV)}, pp. 709--727. Springer, 2022.

\bibitem{reynolds2009gaussian}
D.~A. Reynolds et~al.
\newblock Gaussian mixture models.
\newblock {\em Encyclopedia of biometrics}, 741(659-663), 2009.

\bibitem{Mo_2019_CVPR}
K.~Mo et~al.
\newblock {PartNet}: A Large-Scale Benchmark for Fine-Grained and Hierarchical Part-Level {3D} Object Understanding.
\newblock In {\em Proc. IEEE Conf. Comput. Vis. Pattern Recog. (CVPR)}, 2019.

\bibitem{xue2023zerops}
Y.~Xue et~al.
\newblock ZeroPS: High-quality Cross-modal Knowledge Transfer for Zero-Shot 3D Part Segmentation.
\newblock {\em arXiv preprint arXiv:2311.14262}, 2023.

\bibitem{guo2023sam}
H.~Guo et~al.
\newblock SAM-guided Graph Cut for 3D Instance Segmentation.
\newblock {\em arXiv preprint arXiv:2312.08372}, 2023.

\bibitem{yin2023sai3d}
Y.~Yin et~al.
\newblock SAI3D: Segment Any Instance in 3D Scenes.
\newblock {\em Proc. IEEE Conf. Comput. Vis. Pattern Recog. (CVPR)}, 2024.

\bibitem{xu2023sampro3d}
M.~Xu et~al.
\newblock SAMPro3D: Locating SAM Prompts in 3D for Zero-Shot Scene Segmentation.
\newblock {\em arXiv preprint arXiv:2311.17707}, 2023.

\bibitem{he2024pointseg}
Q.~He et~al.
\newblock PointSeg: A Training-Free Paradigm for 3D Scene Segmentation via Foundation Models.
\newblock {\em arXiv preprint arXiv:2403.06403}, 2024.

\bibitem{peng2023openscene}
S.~Peng et~al.
\newblock Openscene: 3D scene understanding with open vocabularies.
\newblock In {\em Proc. IEEE Conf. Comput. Vis. Pattern Recog. (CVPR)}, pp. 815--824, 2023.

\bibitem{zhang2023clip}
J.~Zhang et~al.
\newblock Clip-fo3d: Learning free open-world 3D scene representations from 2d dense clip.
\newblock {\em Proc. IEEE Int. Conf. Comput. Vis. (ICCV)}, 2023.

\bibitem{ding2022pla}
R.~Ding et~al.
\newblock PLA: Language-Driven Open-Vocabulary 3D Scene Understanding.
\newblock {\em Proc. IEEE Conf. Comput. Vis. Pattern Recog. (CVPR)}, 2023.

\bibitem{yang2024regionplc}
J.~Yang et~al.
\newblock Regionplc: Regional point-language contrastive learning for open-world 3D scene understanding.
\newblock In {\em Proc. IEEE Conf. Comput. Vis. Pattern Recog. (CVPR)}, pp. 19823--19832, 2024.

\bibitem{hasemantic}
H.~Ha and S.~Song.
\newblock Semantic Abstraction: Open-World 3D Scene Understanding from 2D Vision-Language Models.
\newblock In {\em 6th Annual Conference on Robot Learning}, 2022.

\bibitem{takmaz2023openmask3d}
A.~Takmaz et~al.
\newblock Openmask3d: Open-vocabulary 3D instance segmentation.
\newblock {\em Proc. Adv. Neural Inform. Process. Syst. (NeurIPS)}, 2023.

\bibitem{huang2023openins3d}
Z.~Huang et~al.
\newblock Openins3d: Snap and lookup for 3D open-vocabulary instance segmentation.
\newblock {\em Proc. Eur. Conf. Comput. Vis. (ECCV)}, 2024.

\bibitem{lu2023ovir}
S.~Lu et~al.
\newblock Ovir-3D: Open-vocabulary 3D instance retrieval without training on 3D data.
\newblock In {\em Conference on Robot Learning}, pp. 1610--1620, 2023.

\bibitem{nguyen2024open3dis}
P.~Nguyen et~al.
\newblock Open3dis: Open-vocabulary 3D instance segmentation with 2d mask guidance.
\newblock In {\em Proc. IEEE Conf. Comput. Vis. Pattern Recog. (CVPR)}, pp. 4018--4028, 2024.

\bibitem{huang2025segment3d}
R.~Huang et~al.
\newblock Segment3d: Learning fine-grained class-agnostic 3D segmentation without manual labels.
\newblock In {\em Proc. Eur. Conf. Comput. Vis. (ECCV)}, pp. 278--295. Springer, 2025.

\bibitem{boudjoghra2024open}
M.~E.~A. Boudjoghra et~al.
\newblock Open-YOLO 3D: Towards Fast and Accurate Open-Vocabulary 3D Instance Segmentation.
\newblock {\em arXiv preprint arXiv:2406.02548}, 2024.

\bibitem{zhang2023sam3d}
D.~Zhang et~al.
\newblock Sam3d: Zero-shot 3D object detection via segment anything model.
\newblock {\em arXiv preprint arXiv:2306.02245}, 2023.

\bibitem{ding2024vfmm3d}
B.~Ding et~al.
\newblock VFMM3D: Releasing the Potential of Image by Vision Foundation Model for Monocular 3D Object Detection.
\newblock {\em arXiv preprint arXiv:2404.09431}, 2024.

\bibitem{zhang2024fm}
D.~Zhang et~al.
\newblock FM-OV3D: Foundation Model-Based Cross-Modal Knowledge Blending for Open-Vocabulary 3D Detection.
\newblock In {\em Proc. AAAI Conf. Artif. Intell. (AAAI)}, volume~38, pp. 16723--16731, 2024.

\bibitem{panagopoulou2023x}
A.~Panagopoulou et~al.
\newblock X-instructblip: A framework for aligning x-modal instruction-aware representations to llms and emergent cross-modal reasoning.
\newblock {\em arXiv preprint arXiv:2311.18799}, 2023.

\bibitem{qi2024gpt4point}
Z.~Qi et~al.
\newblock Gpt4point: A unified framework for point-language understanding and generation.
\newblock In {\em Proc. IEEE Conf. Comput. Vis. Pattern Recog. (CVPR)}, pp. 26417--26427, 2024.

\bibitem{deitke2024objaverse}
M.~Deitke et~al.
\newblock Objaverse-xl: A universe of 10m+ 3D objects.
\newblock {\em Proc. Adv. Neural Inform. Process. Syst. (NeurIPS)}, 36, 2024.

\bibitem{qi2023contrast}
Z.~Qi et~al.
\newblock Contrast with reconstruct: Contrastive 3D representation learning guided by generative pretraining.
\newblock In {\em Proc. Int. Conf. Mach. Learn. (ICML)}, pp. 28223--28243, 2023.

\bibitem{tang2024minigpt}
Y.~Tang et~al.
\newblock MiniGPT-3D: Efficiently Aligning 3D Point Clouds with Large Language Models using 2D Priors.
\newblock {\em Proc. ACM Int. Conf. Multimedia (ACMMM)}, 2024.

\bibitem{tang2024more}
Y.~Tang et~al.
\newblock More Text, Less Point: Towards 3D Data-Efficient Point-Language Understanding.
\newblock {\em arXiv preprint arXiv:2408.15966}, 2024.

\bibitem{amaduzzi2024llana}
A.~Amaduzzi et~al.
\newblock LLaNA: Large Language and NeRF Assistant.
\newblock {\em arXiv preprint arXiv:2406.11840}, 2024.

\bibitem{mildenhall2021nerf}
B.~Mildenhall et~al.
\newblock Nerf: Representing scenes as neural radiance fields for view synthesis.
\newblock {\em Communications of the ACM}, 65(1):99--106, 2021.

\bibitem{man2024situational}
Y.~Man et~al.
\newblock Situational Awareness Matters in 3D Vision Language Reasoning.
\newblock In {\em Proc. IEEE Conf. Comput. Vis. Pattern Recog. (CVPR)}, pp. 13678--13688, 2024.

\bibitem{hong20233d}
Y.~Hong et~al.
\newblock 3D-LLM: Injecting the 3D world into large language models.
\newblock {\em Proc. Adv. Neural Inform. Process. Syst. (NeurIPS)}, 2023.

\bibitem{cheng2024spatialrgpt}
A.-C. Cheng et~al.
\newblock SpatialRGPT: Grounded Spatial Reasoning in Vision Language Model.
\newblock {\em Proc. Adv. Neural Inform. Process. Syst. (NeurIPS)}, 2024.

\bibitem{liu2024uni3d}
D.~Liu et~al.
\newblock Uni3d-LLM: Unifying point cloud perception, generation and editing with large language models.
\newblock {\em arXiv preprint arXiv:2402.03327}, 2024.

\bibitem{lin2023sphinx}
Z.~Lin et~al.
\newblock Sphinx: The joint mixing of weights, tasks, and visual embeddings for multi-modal large language models.
\newblock {\em arXiv preprint arXiv:2311.07575}, 2023.

\bibitem{zhu2024llava}
C.~Zhu et~al.
\newblock LLaVA-3D: A Simple yet Effective Pathway to Empowering LMMs with 3D-awareness.
\newblock {\em arXiv preprint arXiv:2409.18125}, 2024.

\bibitem{kerbl3Dgaussians}
B.~Kerbl et~al.
\newblock 3D Gaussian Splatting for Real-Time Radiance Field Rendering.
\newblock {\em ACM Transactions on Graphics}, 42(4), 2023.

\bibitem{objaverse}
M.~Deitke et~al.
\newblock Objaverse: A Universe of Annotated 3D Objects.
\newblock In {\em Proceedings of the IEEE/CVF Conference on Computer Vision and Pattern Recognition}, pp. 13142--13153, 2023.

\bibitem{ye2023closer}
C.~Ye et~al.
\newblock A closer look at few-shot 3d point cloud classification.
\newblock {\em International Journal of Computer Vision}, 131(3):772--795, 2023.

\bibitem{ahmadi2024foundation}
S.~Ahmadi et~al.
\newblock Foundation Model-Powered 3D Few-Shot Class Incremental Learning via Training-free Adaptor.
\newblock In {\em Proceedings of the Asian Conference on Computer Vision}, pp. 2282--2299, 2024.

\bibitem{yang2023geometry}
Y.~Yang et~al.
\newblock Geometry and Uncertainty-Aware 3D Point Cloud Class-Incremental Semantic Segmentation.
\newblock In {\em Proceedings of the IEEE/CVF Conference on Computer Vision and Pattern Recognition (CVPR)}, pp. 21759--21768, June 2023.

\bibitem{lomonaco2020continual}
V.~Lomonaco et~al.
\newblock Continual reinforcement learning in 3d non-stationary environments.
\newblock In {\em Proceedings of the IEEE/CVF Conference on Computer Vision and Pattern Recognition Workshops}, pp. 248--249, 2020.

\end{thebibliography}
